\pdfoutput=1
\documentclass[sigconf, nonacm]{acmart}
\usepackage{tikz}
\usepackage{pgfplots}
\usepackage{subfigure}
\usepackage{graphicx}
\setcopyright{none}
\pgfplotsset{compat = 1.3}
\AtBeginDocument{%
  \providecommand\BibTeX{{%
    \normalfont B\kern-0.5em{\scshape i\kern-0.25em b}\kern-0.8em\TeX}}}

\setcopyright{acmcopyright}
\copyrightyear{2018}
\acmYear{2018}
\acmDOI{10.1145/1122445.1122456}

\acmConference[Woodstock '18]{Woodstock '18: ACM Symposium on Neural
  Gaze Detection}{June 03--05, 2018}{Woodstock, NY}
\acmBooktitle{Woodstock '18: ACM Symposium on Neural Gaze Detection,
  June 03--05, 2018, Woodstock, NY}
\acmPrice{15.00}
\acmISBN{978-1-4503-XXXX-X/18/06}

\begin{document}

\title{Embracing the Dark Knowledge: Domain Generalization Using Regularized Knowledge Distillation}


\author{Yufei Wang}
\affiliation{%
  \institution{Rapid-Rich Object Search Lab,\\Nanyang Technological University}
  \country{}}
\email{im.wangyufei@gmail.com}

\author{Haoliang Li}
\authornote{Corresponding author}
\affiliation{%
  \institution{Department of Electrical Engineering,\\City University of Hong Kong}
  \country{}}
\email{haoliang.li@cityu.edu.hk}

\author{Lap-pui Chau}
\affiliation{%
  \institution{Rapid-Rich Object Search Lab,\\Nanyang Technological University}
  \country{}}
\email{elpchau@ntu.edu.sg}

\author{Alex C. Kot}
\affiliation{%
  \institution{Rapid-Rich Object Search Lab,\\Nanyang Technological University}
  \country{}}
\email{eackot@ntu.edu.sg}







\begin{abstract}
  Though convolutional neural networks are widely used in different tasks, lack of generalization capability in the absence of sufficient and representative data is one of the challenges that hinders their practical application. In this paper, we propose a simple, effective, and plug-and-play training strategy named \underline{K}nowledge \underline{D}istillation for \underline{D}omain \underline{G}eneralization (KDDG) which is built upon a knowledge distillation framework with the gradient filter as a novel regularization term. We find that both the ``richer dark knowledge" from the teacher network, as well as the gradient filter we proposed, can reduce the difficulty of learning the mapping which further improves the generalization ability of the model. We also conduct experiments extensively to show that our framework can significantly improve the generalization capability of deep neural networks in different tasks including image classification, segmentation, reinforcement learning by comparing our method with existing state-of-the-art domain generalization techniques.  Last but not the least, we propose to adopt two metrics to analyze our proposed method in order to better understand how our proposed method benefits the generalization capability of deep neural networks.
\end{abstract}

\ccsdesc[500]{Computing methodologies~Computer vision}



\keywords{domain generalization, knowledge distillation, robustness}


\maketitle

\section{Introduction}
Recently, convolution neural networks are widely used in different tasks and scenarios thanks to the development of machine learning and computing devices. However, one of the challenges that hinders the practical application of the convolution neural network model is the lack of generalization capability in the absence of sufficient and representative data \cite{zhang2016understanding, kawaguchi2017generalization}. To overcome such limitation, domain adaptation (DA) have been widely studied under the assumption that a small amount of labeled target domain data or unlabeled target domain data can be obtained during the training stage.
\begin{figure}[tbp]
\begin{center}
\includegraphics[width=1\linewidth, trim=20 15 60 20, clip]{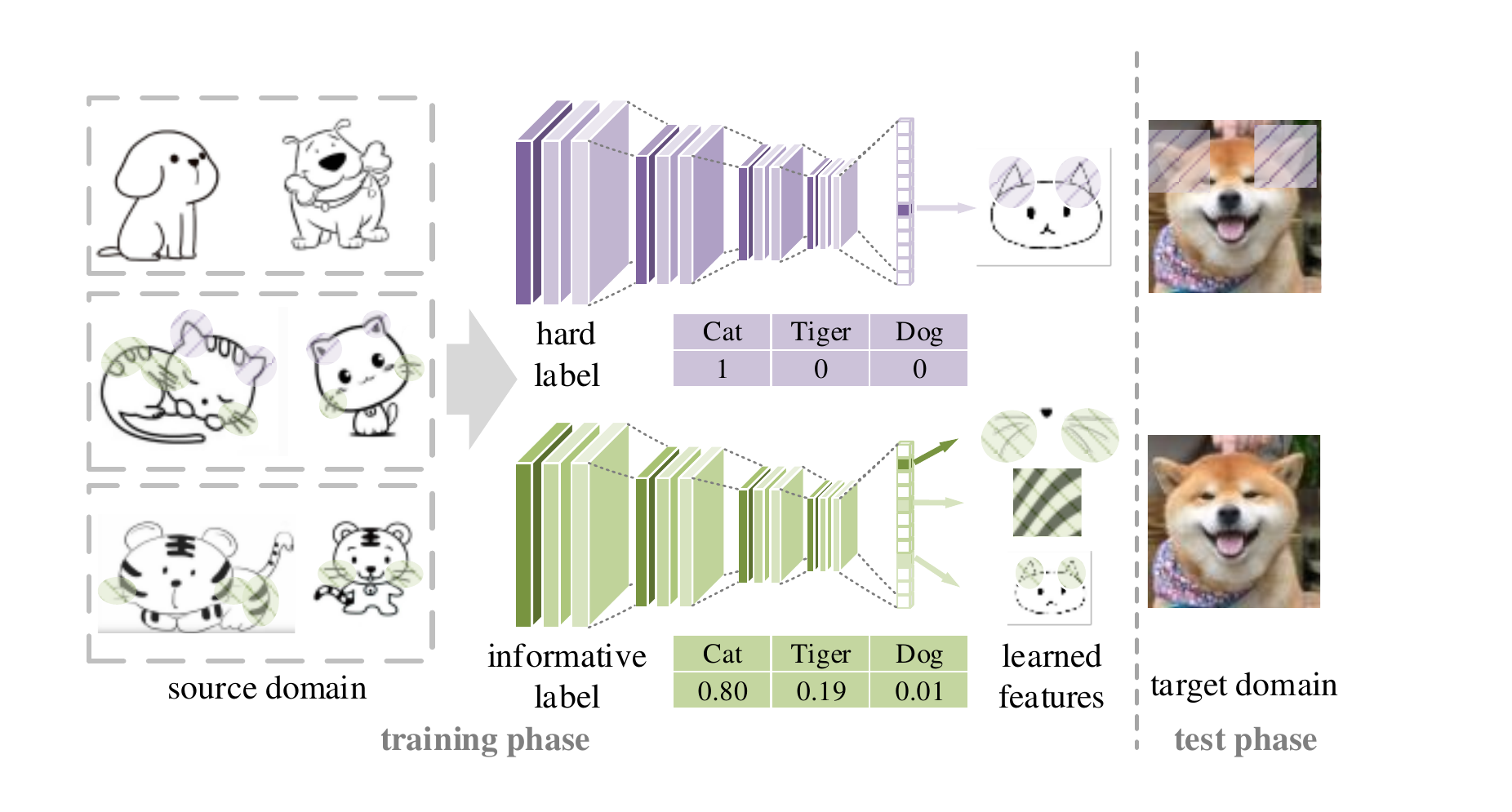}
\end{center}
   \caption{A motivating example of cat recognition. If the source domain data is biased, e.g., only the cat’s ears are triangular, the model using hard labels will easily overfit to the feature of triangle ear which is not domain invariant. However, if the student model uses the informative soft label from the teacher, it has to learn more general features such as the whiskers and stripes that are shared between cats and tigers and then predict the category using the combination of these features. For a dog image from the unseen domain also with triangle ears, the network trained with the hard labels may be wrongly activated.}
\label{fig:insight}
\vspace{-0.1cm}
\end{figure}

\begin{figure}[tbp]
    \centering
\subfigure[cartoon]{
\scalebox{1}{
    \begin{minipage}[b]{1.60cm}
    \includegraphics[height=1.60cm, width=1.60cm]{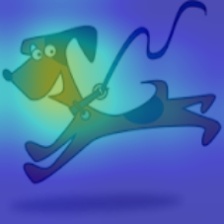} \\
    \includegraphics[height=1.60cm, width=1.60cm]{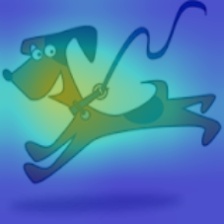} 
    \end{minipage}
    }
}
\hspace{-4.5mm}
\subfigure[sketch]{
\scalebox{1}{
    \begin{minipage}[b]{1.60cm}
    \includegraphics[height=1.60cm, width=1.60cm]{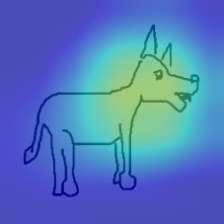} \\
    \includegraphics[height=1.60cm, width=1.60cm]{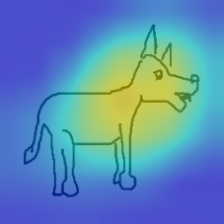} 
    \end{minipage}
    }
}
\hspace{-4.5mm}
\subfigure[photo]{
\scalebox{1}{
    \begin{minipage}[b]{1.60cm}
    \includegraphics[height=1.60cm, width=1.60cm]{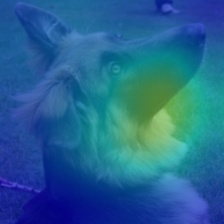} \\
    \includegraphics[height=1.60cm, width=1.60cm]{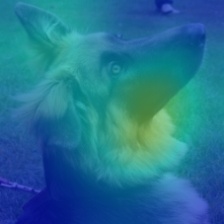} 
    \end{minipage}
    }
}
\hspace{-4.5mm}
\subfigure[\textbf{painting}]{
\scalebox{1}{
    \begin{minipage}[b]{1.60cm}
    \includegraphics[height=1.60cm, width=1.60cm]{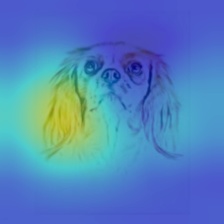} \\
    \includegraphics[height=1.60cm, width=1.60cm]{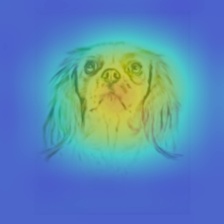} 
    \end{minipage}
    }
}
    
    \caption{The visualization of the class activation map \cite{zhou2016learning} from source and target domains using PACS benchmark. The first row is the baseline method DeepAll and the second row is our method KDDG. The first three columns are source domains and the last is the unseen target domain.}
    \label{fig:CAM}
\end{figure}

However, in some cases, the target domain data may not always be available. As such, directly training on the source domain can lead to an overfitting problem. In other words, we can observe a significant performance drop when testing on the target domain. Domain generalization (DG) is one research direction that aims to utilize multi-source domains to train a model that is expected to be generalized better on the unseen target domains. The core idea of existing methods are either to align the features of source domains to a pre-defined simple distribution, to learn a domain invariant feature distribution \cite{yang2013multi, muandet2013domain, li2018domain}, or to conduct data augmentation (e.g., data enhancement \cite{wang2020heterogeneous, zhang2019unseen}, generative adversarial network (GAN) based methods \cite{volpi2018generalizing,zhou2020deep}) on source domains. In addition, there are also some efforts using the meta-learning based method to simulate the domain gap between the source and unseen target domain. However, the aforementioned techniques inevitably face the problem of overfitting on the source domain due to the inability to access the target domain data. It is still an open problem on how to learn a universal feature representation through the aforementioned techniques that the latent features of target lie on a similar feature distribution compared with the latent features of source domains. As such, the generalization performance in the target domain may not be guaranteed due to the distribution mismatch. 
\begin{figure}[tbp]
\begin{center}
\scalebox{0.8}{
    \centering
%
%
\definecolor{mycolor1}{rgb}{0.00000,0.44700,0.74100}%
\definecolor{mycolor2}{rgb}{0.85000,0.32500,0.09800}%
\begin{tikzpicture}
\begin{axis}[%
width=2.343in,
height=1.436in,
at={(-0.35in,0in)},
scale only axis,
xmin=0,
xmax=0.35,
xlabel style={font=\color{white!15!black}},
xlabel={the ratio $\lambda$ of the noisy labels},
ymin=0.6,
ymax=1,
ylabel style={font=\color{white!15!black}},
ylabel={average accuracy},
axis background/.style={fill=white},
legend style={legend cell align=left, align=left, draw=white!15!black}
]
\addplot [color=mycolor1, line width=1.8pt, mark=o, mark options={solid, mycolor1}]
  table[row sep=crcr]{%
0	0.9545\\
0.05	0.9428\\
0.1	0.9303\\
0.15	0.9153\\
0.2	0.9039\\
0.25	0.885\\
0.3	0.8627\\
0.35	0.8484\\
};
\addlegendentry{source}

\addplot [color=mycolor2, dashed, line width=1.8pt, mark=o, mark options={solid, mycolor2}]
  table[row sep=crcr]{%
0	0.88897074\\
0.05	0.84153368\\
0.1	0.81997202\\
0.15	0.77833899\\
0.2	0.74335283\\
0.25	0.71043051\\
0.3	0.67073748\\
0.35	0.638204\\
};
\addlegendentry{target}

\end{axis}

\begin{axis}[%
width=2.343in,
height=1.536in,
at={(0in,0in)},
scale only axis,
xmin=0,
xmax=1,
ymin=0,
ymax=1,
axis line style={draw=none},
ticks=none,
axis x line*=bottom,
axis y line*=left
]
\end{axis}
\end{tikzpicture}%
}
\end{center}
   \caption{The impact of task difficulty on the generalization performance on the source and unseen target domains. We change the ratio $\lambda$ of the noise label (the label is fixed when start training) to change the difficulty of the task \cite{zhang2016understanding, nettleton2010study}. The experiments were done using Digit-Five  \cite{peng2019moment} benchmark and we set syn as the target domain. {The network is weak to predict the randomly initialized labels which are inconsistent with the inductive preference of the model, i.e., the recognition task becomes difficult when increasing $\lambda$.}}
\label{fig:noise-level}
\end{figure}
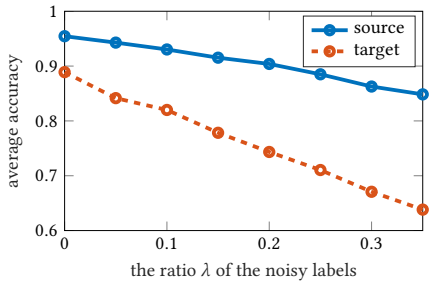
Our motivation comes from the observation that if the task is difficult for the model, i.e., the mapping from the input space to the label space is difficult to learn, the performance of the model in the target domain may have a large drop owing to the overfitting in source domains even if the performance on the source domains is still acceptable. One illustrative example to verify our observation can be seen in Fig. \ref{fig:noise-level}. By involving more noisy labels on the source domain, the learning task of the model also becomes difficult \cite{zhang2016understanding, nettleton2010study} and we find it further deteriorates the performance of the model on the target domain. To this end, we make the assumption that the generalization performance of the model in the unseen target domain can benefit from the easy tasks. 
In particular, we proposed a framework named knowledge distillation for domain generalization (KDDG) with a novel gradient regularization to decrease the difficulty of the task. The motivation is mainly in two folds. 1) we leverage the advantage of knowledge distillation to provide the student with more informative labels which make the training easier \cite{phuong2019towards, hinton2015distilling}, where the rationales are illustrated in Fig. \ref{fig:insight}. We also provide a preliminary study in Fig. \ref{fig:CAM} using class activation map (CAM) \cite{zhou2016learning}. We observe that our proposed method can help the student network to learn more general and comprehensive features, e.g., the legs and the body of the dog which are ignored by directly training the model with one-hot ground truth labels.  2) the gradient filter we proposed relaxes the objective of IRM  that further makes the task easier (which is analyzed in Sec. \ref{sec:GradFilterDiscus} in details) and avoid the student model from completely imitating the teacher \cite{phuong2019towards}, such that better generalization can be further guaranteed. We show that our proposed method can outperform existing state-of-the-art domain generalization techniques on different tasks including image classification, segmentation, reinforcement learning. We further propose to adopt two different metrics, namely cumulative weight distance and mutual information, to explain the effectiveness of our proposed method. The main idea is to measure the difficulty of the task and quantify the domain-specific information. We show that our proposed method KDDG can make the task easier to train and allow the student network to learn fewer domain-specific features, which further justify our assumption to tackle the problem of domain generalization through knowledge distillation. In summary, we make the following contributions,
\begin{enumerate}
  \item We propose to tackle the problem of domain generalization from a perspective of task difficulty and reveal that better generalization capability can be achieved if the learning task is easier. 
  \item We propose a simple, effective, and plug-and-play training strategy named KDDG. Experimental results on different tasks demonstrate the effectiveness of our proposed method by comparing it with the SOTA techniques. 
  \item Two metrics are proposed to better understand how our proposed method benefits the generalization capability of neural networks.
\end{enumerate}

\section{Related Work}
\subsection{Domain Adaptation and Generalization}
How to alleviate the performance degradation caused by the domain gap between the training and test data has always been a hot research pot. One direction is to rely on unlabeled test data (a.k.a., target domain data) and apply domain adaptation (DA) or transfer learning \cite{huang2006correcting, pan2009survey, li2020unsupervised}, which can be roughly divided into two categories, namely subspace learning based methods and instance re-weighting \cite{huang2006correcting,pan2011domain,zhang2015multi,ghifary2017scatter}. In addition, deep learning based methods also demonstrate their effectiveness by feature alignment using different methods such as Maximum Mean Discrepancy (MMD) \cite{long2015learning} and adversarial based training \cite{ganin2016domain, tzeng2017adversarial, bousmalis2017unsupervised, hoffman2017cycada}.

The setting of domain generalization (DG) is similar but more challenging than the setting of domain adaptation because we cannot get any information from the target domain in the training phase. Part of the current DG methods borrow the idea from DA, e.g., domain invariant feature and feature alignment. For example,  for domain invariant feature-based methods, \cite{yang2013multi} aimed to use Canonical Correlation Analysis (CCA) to obtain the shareable information, \cite{muandet2013domain} proposed a domain invariant analysis method which also used MMD and was further extended by \cite{li2018domain, li2020gmfad}. Multi-task auto-encoder was also used to learn shareable feature representations \cite{ghifary2015domain}. In addition, different regularization methods were proposed, e.g., low-rank regularization was used in \cite{xu2014exploiting,li2017deeper} to extract the invariant embedding, the puzzle task \cite{carlucci2019domain} was used to learn a model with good generalization performance. Meta-based methods \cite{li2018learning, balaji2018metareg} can be also treated as a kind of regularization, e.g., \cite{li2018learning} used the second derivative as a regular term and \cite{balaji2018metareg} directly learned a regularization network. However, we may still not be able to guarantee that the embedding from target domains share the same subspace with that of the source domains through the aforementioned domain invariant feature learning based methods since the learned classifier can be overfitted to the source domains. In addition, the regularization may make the learning task difficult which further increases the risk of overfitting in the source domains. There also exists some works using data augmentation based techniques \cite{volpi2018generalizing, zhou2020deep,wang2020heterogeneous}. For example,  \cite{zhou2020deep} used an adversarial generative model to generate some samples that are out of the source domains 
to improve the generalization capability of the neural network. However, if the distribution of the newly generated data is not similar to that of the target domain, it may have a negative impact on the model. In addition, it may be difficult to generate samples out of the convex set of the source domains \cite{berthelot2018understanding}.
\subsection{Knowledge Distillation}
To the best of our knowledge, knowledge distillation (KD) was first designed for model compression \cite{bucilu2006model, sau2016deep},  which aims to make the output of the new model (student) similar to that of the previous model (teacher model) meanwhile decrease the size of the student one. In \cite{hinton2015distilling, luo2016face}, it has been shown that training the student model with ``dark knowledge" by KD can lead to better performance compared with directly training with one-hot ground truth. Besides, it has also been shown that KD can benefit nature language processing \cite{li2014learning, zhao2019extreme}. In the area of reinforcement learning, knowledge distillation is also known as policy distillation, which can be applied for model compression, network training acceleration, or multiple agent models merging \cite{rusu2015policy, parisotto2015actor, yin2017knowledge}. Comparing with the application of KD, the theoretical analysis \cite{phuong2019towards, lopez2015unifying, cheng2020explaining, muller2019does} is relatively deficient and mainly focus on the characteristics such as the convergence \cite{phuong2019towards}.  While there exist some works linking KD with the generalization capability of neural network (e.g., \cite{phuong2019towards,muller2019does}), they mainly focus on analyzing the generalization capability of the model using the same dataset or data distribution, e.g., the upper bound of the transfer risk \cite{phuong2019towards}. Different from the previous work, we explore and analyze KD from a new perspective by focusing on the setting of domain generalization where the testing data are collected from a different distribution compared with the training data.
 


\section{Methodology}
\subsection{Knowledge Distillation}
Before introducing our proposed method, we first revisit the idea of knowledge distillation \cite{hinton2015distilling}. Specifically, in order to obtain the soft label, temperature $\tau$ is introduced as a hyper-parameters to soften the vanilla softmax distribution, which is given as
\begin{equation}
    p^i (x;\tau) = softmax(s(x);\tau)=\frac{e^{s_i(x)/\tau}}{\sum_k e^{s_k(x)/\tau}}
\end{equation}
where $s_i(x)$ is the score logit from the sample $x$ of class $i$.
Subsequently, the KL-divergence is used as the distillation loss $L_{kd}$ to measure the difference between the teacher and student:
\begin{equation}
    L_{kd} = -\tau ^ 2 \mathbb{E} _{x\sim D_{s}}\sum_{i=1}^{C}p_t^i(x;\tau)log(p_s^i(x;\tau))
\end{equation}
where $p_t$ and $p_s$ denote the softened probability from the teacher and student network respectively, $C$ is the total number of the categories, $D_s$ indicates the distribution of all source domain data by concatenating them together. Here, the coefficient $\tau^2$ is used to balance the magnitude of the gradient \cite{hinton2015distilling} between hard and soft label. The loss function for the student network is as follows:
\begin{equation}
    L_{vanilla}= \lambda_1 L_{ce} + \lambda_2 L_{kd}
\label{student_loss}
\end{equation}
where $L_{ce}$ is the standard cross-entropy, $\lambda_1$ and $\lambda_2$ are balancing weight. 

\subsection{Gradient Filter}
\label{sec:GradFilter}
We further propose a novel gradient regularization to improve the generalization capability of the network.
Without loss of generality, we use the classification task to illustrate the gradient filter $f$. Assuming that $p(x_i)$ denote the softmax output vector of the student network from the sample $x_i$, $p^k(x_i)$ represents the probability value of the sample $x_i$ with respect to its ground truth category $k$ (abbreviated as $p$). We can define the gradient filter $f$ as
\begin{equation}
f(\omega) = \left\{\begin{matrix}
\omega & p \leq \eta \\ 
(\frac{\eta+1-2p}{1-\eta})^2 \omega &\eta < p \leq \frac{1+\eta}{2}\\
0 & p > \frac{1+\eta}{2} \\
\end{matrix}\right.
\label{gradient_filter}
\end{equation}
where $\omega$ denotes gradient and $\eta$ is a hyper-parameter that controls the intensity of the filter. Noted that we can also apply the operation $f$ on the empirical risk of a sample, as our proposed GradFilter can be treated as imposing weight on the loss function. 

The significance of $f$ comes from mainly two folds, 1) we can prevent the student network from being too similar to the teacher network \cite{phuong2019towards}, such that the risk of overfitting to teacher network can be reduced; 2) by filtering out the gradient corresponding to high score output, we can avoid the problem of over-confidence \cite{guo2017calibration,muller2019does} and make the optimization task easier by relaxing the objective of IRM which will be further discussed in Sec. \ref{sec:GradFilterDiscus}.

\subsection{Knowledge Distillation for DG}

Now we introduce our proposed method built upon knowledge distillation with gradient filter regularization. Specifically, at each iteration, the gradient filter inspects the confidence of each sample and decrease the gradient weight of the sample which has a confidence score higher than a pre-defined threshold (by setting the corresponding loss to zero). Thus, the integrated loss for domain generalization can be formulated as
\begin{equation}
    L_{kd}^{f} = -\tau ^ 2 \mathbb{E}_{x\sim D_{s}}\epsilon (f(\sum_{i=1}^{C}p_t^i(x;\tau)log(p_s^i(x;\tau))))
\label{kd_loss_with_filter}
\end{equation}
\begin{equation}
    L_{ce}^{f} = -\mathbb{E}_{x\sim D_{s}}f(\sum_{i=1}^{C}(y^i log(p_s^i(x)))
\end{equation}
where $y^i$ denotes the ground truth of the sample $x$ for class $i$. $\epsilon$ is defined as follows to avoid the bad effect from the wrong predict from the teacher
\begin{equation}
    \epsilon(l) = 
\left\{\begin{matrix}
l & \upsilon(p_{t}(x_i)) = y_i \\ 
0 & \text{otherwise}
\end{matrix}\right.
\end{equation}
where $p^k_s(x_i)$ and $p_t$ represent the probability value of the sample $x_i$ with the class index $k$ from the student network and probability vector from the teacher network respectively, $\upsilon$ is an operation that returns the predicted label according to the probability vector. The final objective for KDDG can be represented as 
\begin{equation}
    L_{KDDG} = \lambda_1 L_{kd}^{f}+ \lambda_2 L_{ce}^{f}.
\end{equation}
For KD, we set $\tau$ to 2. We also set $\lambda_1$ and $\lambda_2$ to 0.5
so that we have roughly the same magnitude of the gradient for fair comparison and easy analysis.


\subsection{Discussion}
\label{sec:GradFilterDiscus}
To further understand the rationale and effectiveness of gradient filter we proposed, we give a view from the perspective of invariant risk minimization (IRM) \cite{ahuja2020invariant, arjovsky2019invariant}, which is for generalization analysis of the classifier across multiple domains.


According to IRM, the task of DG can be treated as to find a representation $\Phi$ such that the optimal classifier given $\Phi$ is invariant across different domains. More specifically, for a feature extractor $\Phi$, there exists an invariant predictor $w$ if this predictor achieves the best performance among all the domains. In \cite{ahuja2020invariant, arjovsky2019invariant}, the authors proposed to find the optimal $\Phi$ and $w$ as
\begin{equation}
\begin{aligned}
    & \underset{\Phi \in \mathcal{H}_\Phi, w\in\mathcal{H}_w}{\min} \underset{d \in \mathcal{D}}{\sum} \mathcal{L}^d(w\circ \Phi) \\
    & \text{ s.t. } w\in \arg \underset{\bar{w}\in \mathcal{H}_w}{\min } \mathcal{L}^d(\bar{w}\circ \Phi), \forall d \in \mathcal{D}  
\label{invariant}
\end{aligned}
\end{equation}
where $\mathcal{L}^d$ is the empirical risk in the domain $d$ and $\mathcal{D}$ represents all the domains, and the solution space of $(w, \Phi)$ can be defined as $\mathcal{S}^{IV}$.


Unlike the previous works about IRM \cite{ahuja2020invariant, arjovsky2019invariant}, our GradFilter can be treated as to conduct solution space transformation on $\mathcal{S}^{IV}$. By applying GradFilter operation, Eq. \ref{invariant} can be reformulated as
\begin{equation}
\begin{aligned}
    & 
\underset{\hat{\Phi} \in \mathcal{H}_\Phi, \hat{w}\in\mathcal{H}_w}
{\min}
\underset{d \in \mathcal{D}}{\sum} f (\mathcal{L}^d(\hat{w}\circ \hat{\Phi}))
\\
    & \text{ s.t. } \hat{w}\in \arg \underset{\tilde{w}\in \mathcal{H}_w}{\min }f(\mathcal{L}^d(\tilde{w}\circ \Phi)), \forall d \in \mathcal{D}
\label{invariant_gradientFilter}
\end{aligned}
\end{equation}
where $f$ is the GradFilter operation defined in Eq. \ref{gradient_filter} which avoid the model over-confidence in the source domains.

Our proposed GradFilter can be interpreted as to relax the objective of IRM. If directly optimizing Eq. \ref{invariant}, it may be hard to find a solution which satisfies both the objective and the regularization term, as there may not exist an intersection among the solution set of each source domain. By relaxing the objective, we make the optimization task easier, i.e., it is more likely that there exists an overlap region among solution space between source domains. Thus, our proposed method can be interpreted as an alternative solution for Eq. \ref{invariant}. More details can be found in the supplementary materials.

\begin{figure*}[htbp]
\centering
\scalebox{1.05}{
\subfigure[Digit-Five\cite{peng2019moment}]{
    \scalebox{0.6}{
    \begin{minipage}[b]{0.6cm}
    \includegraphics[height=0.6cm, width=0.6cm]{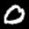} \\
    \includegraphics[height=0.6cm, width=0.6cm]{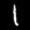} \\
    \includegraphics[height=0.6cm, width=0.6cm]{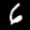} \\
    \includegraphics[height=0.6cm, width=0.6cm]{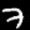}
    \end{minipage}
    \begin{minipage}[b]{0.6cm}
    \includegraphics[height=0.6cm, width=0.6cm]{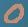} \\
    \includegraphics[height=0.6cm, width=0.6cm]{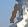} \\
    \includegraphics[height=0.6cm, width=0.6cm]{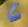} \\
    \includegraphics[height=0.6cm, width=0.6cm]{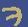} 
    \end{minipage}
    \begin{minipage}[b]{0.6cm}
    \includegraphics[height=0.6cm, width=0.6cm]{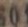} \\
    \includegraphics[height=0.6cm, width=0.6cm]{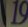} \\
    \includegraphics[height=0.6cm, width=0.6cm]{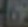} \\
    \includegraphics[height=0.6cm, width=0.6cm]{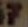} 
    \end{minipage}
    \begin{minipage}[b]{0.6cm}
    \includegraphics[height=0.6cm, width=0.6cm]{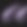} \\
    \includegraphics[height=0.6cm, width=0.6cm]{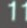} \\
    \includegraphics[height=0.6cm, width=0.6cm]{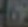} \\
    \includegraphics[height=0.6cm, width=0.6cm]{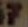} 
    \end{minipage}
    \begin{minipage}[b]{0.6cm}
    \includegraphics[height=0.6cm, width=0.6cm]{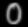} \\
    \includegraphics[height=0.6cm, width=0.6cm]{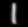} \\
    \includegraphics[height=0.6cm, width=0.6cm]{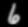} \\
    \includegraphics[height=0.6cm, width=0.6cm]{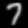} 
    \end{minipage}
    }
}
\hspace{-3mm}
\subfigure[mini-DomainNet \cite{peng2019moment,zhou2020domain}]{
\includegraphics[height=1.5cm]{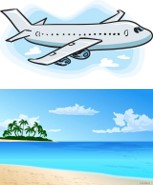} \hspace{-1mm}
\includegraphics[height=1.5cm, trim=0 0 0 0, clip]{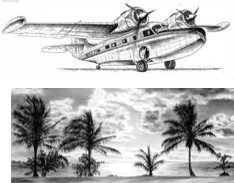}
\includegraphics[height=1.5cm]{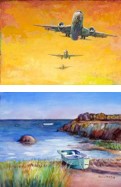}
\includegraphics[height=1.5cm]{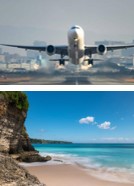}
}
\hspace{-1mm}
\subfigure[Mountain-Car \cite{brockman2016openai}]{
\includegraphics[height=1.5cm]{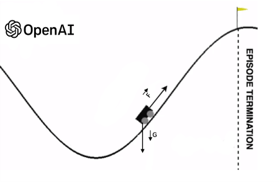}
\label{fig:mountain_car}
}
\hspace{-3mm}
\subfigure[PACS \cite{li2017deeper}]{
\includegraphics[height=1.5cm]{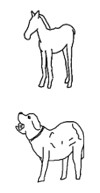}
\includegraphics[height=1.5cm]{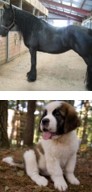}
\includegraphics[height=1.5cm]{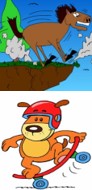}
\includegraphics[height=1.5cm]{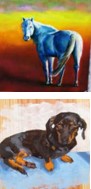}
}
\hspace{-2mm}
\subfigure[gray matter segmentation \cite{prados2017spinal}]{
\includegraphics[height=1.5cm]{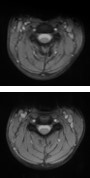}  
\includegraphics[height=1.5cm]{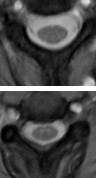}  
\includegraphics[height=1.5cm]{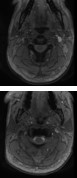}  
\includegraphics[height=1.5cm]{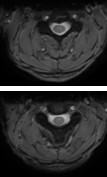} \hspace{1mm}
\label{fig:gray_matter}
}}
\caption{An illustration of domain generalization benchmarks used for evaluation. Each column represents a domain and the data are randomly sampled except the reinforcement learning task Mountain-Car where the domain gap is concentrated in the different gravitational constant.}
\label{fig:benchmark}\end{figure*}

\section{Experiments}
We evaluate our methods on different domain generalization benchmarks, including Digit-Five \cite{peng2019moment}, PACS \cite{li2017deeper}, mini-domainnet \cite{peng2019moment,zhou2020domain}, gray matter segmentation task \cite{prados2017spinal} and mountain car \cite{brockman2016openai}. Examples of each benchmark are shown in Fig. \ref{fig:benchmark}. {For all the experiments, we use the same architecture for both teacher and student network. In addition, the gradient filter is only used when training the student network based on its output.} Due to the limited space, more details about the experiments are in the supplementary material.
\subsection{Digits-Five}
Digits-Five is a benchmark used by \cite{peng2019moment}, which is composed of MNIST-M, MNIST, SVHN, USPS and SYN with different backgrounds, fonts, styles, etc.

\textbf{Settings:} We follow the experiment settings in \cite{peng2019moment} except that we assume the target domain is unavailable during training. More specifically, we adopt a standard leave-one-domain-out manner. For the source domains, 80\% of the samples are used for training and 20\% for validation. All the samples in the target domain are used for testing. We follow \cite{peng2019moment} to convert all image samples with the resolution of $32\times32$ in RGB format.  We use the same backbone network for all methods which include three convolutional blocks and two FC layers. Each convolutional block contains a $3\times3$ convolutional layer, a batch normalization layer, a Relu layer, and a $2\times2$ max-pooling layer. The SGD is used with an initial learning rate of 0.05 and a weight decay of 5e-4 for 30 epochs. The cosine annealing scheduler is used to decrease the learning rate.

\textbf{Results:} We compare our methods with several competitive models including MLDG \cite{li2018learning}, JiGen \cite{carlucci2019domain}, MASF \cite{dou2019domain} and RSC \cite{huang2020self}. We report the baseline results in Table \ref{tab:digit5} by tuning the hyper-parameters in a wide range. As we can observe, our method achieves the best results which prove the effectiveness of our proposed method. In addition, using the knowledge distillation alone also has some improvement which further verifies our motivation.

\begin{table}[htbp]
    \centering
    \scalebox{1}{
    \begin{tabular}{ccccccc}
    \toprule
        & MM & MNIST & USPS & SVHN & SYN & Avg.  \\
    \midrule
        DeepAll & 66.3 & 96.7 & 94.1 & 82.1 & 88.9 & 85.6 \\
        MLDG \cite{li2018learning} & 67.9 & 97.3 & 94.7	& 83.9 & 89.1 & 86.6 \\
        JiGen \cite{carlucci2019domain} & 67.8 & 97.8 & 95.9 & 84.7 & 89.6 & 87.2 \\
        MASF \cite{dou2019domain} & 69.2 & 98.6 & 96.3 & 84.4 & 90.1 & 87.7 \\
        RSC \cite{huang2020self} & 69.8	& 98.3 & 96.1 & 84.1 & 89.9 & 87.6 \\
        \hline
        Distill & 69.7 & \textbf{99.2} & 97.3 & 84.7& 90.4 & 88.3\\
        KDDG(Ours) & \textbf{70.5} & 99.1 & \textbf{97.6} & \textbf{85.5} & \textbf{90.6} & \textbf{88.7} \\
    \bottomrule
    \end{tabular}
    }
    \caption{Evaluation of DG on the Digit-Five benchmark. The average target domain accuracy of five repeated experiments is reported. \emph{MM} and \emph{SYN} are abbreviations for \emph{MNIST-M} and \emph{Synthetic Digits}.}
    \label{tab:digit5}
\end{table}

\subsection{PACS}
\label{sec:PACS}
PACS \cite{li2017deeper} is a standard benchmark for domain generalization which includes 4 different domains: photo, sketch, cartoon, painting. There are 7 categories in the dataset and 9991 images in total.

\textbf{Settings:} Following the experiment settings in \cite{carlucci2019domain}, we use the ImageNet pre-trained ResNet18 as the backbone network for all methods. The samples from source domains are divided into training (90\%) and validation (10\%) using the official train-val splits. All the images from the target domain are used for test. SGD is used to train all the networks with an initial learning rate of 5e-4, batch size of 16, and weight decay 5e-4 for 25 epochs. The learning rate is decreased to 5e-5 at the $20$th epoch. It is worth noting that we use the same data argumentation for all methods which include random crop on images with a scale factor of 1.25 and random horizontal flip. We do not use the augmentations, e.g., random gray scale, considering that it can introduce the prior knowledge of specific domain information, e.g., the samples in the sketch domain are all grayscale which are not available in the DG setting. 

\textbf{Results:} To compare the performance of different methods, we report the top 1 classification accuracy in the unseen target domain. We repeat the experiment for 5 times and report the average target domain accuracy at the last epoch. Several state of the art methods are used for comparison, including CCSA \cite{yoon2019generalizable}, MLDG \cite{li2018learning}, CrossGrad \cite{shankar2018generalizing}, eta. The results are shown in Table \ref{tab:pacs}. {It is worth noting that our method is different from MASF \cite{dou2019domain} which minimizes the divergence of the class-specific mean feature vectors between meta-train and meta-test domains. Specifically, we consider to conduct knowledge distillation in an one-one correspondence manner based on the same input between teacher and student network, which may avoid negative transfer and we also empirically find that if can lead to better performance.} 
\begin{table}[]
    \centering
    \scalebox{1}{
    \begin{tabular}{cccccc}
    \toprule
        & Art & Cartoon & Photo & Sketch & Avg.  \\
    \midrule
        DeepAll & 77.0 & 75.9 & 95.5 & 70.3 & 79.5\\ 
        CCSA \cite{yoon2019generalizable}& 80.5 & 76.9 & 93.6 & 66.8 & 79.4\\ 
        MLDG \cite{li2018learning}& 78.9 & 77.8 & 95.7 & 70.4 & 80.7 \\
        CrossGrad \cite{shankar2018generalizing} &79.8 & 76.8 & \textbf{96.0} & 70.2 & 80.7\\
        JiGen \cite{carlucci2019domain} & 79.4 & 75.3 & \textbf{96.0} & 71.6 & 80.5\\
        MASF \cite{dou2019domain} & 80.3 & 77.2 & 93.9 & 71.7 & 81.0\\
        Epi-FCR \cite{li2019episodic} & \textbf{82.1} & 77.0 & 93.9 & \textbf{73.0} & 81.5\\
        RSC \cite{huang2020self} & 79.5 & 77.8 & 95.6 & 72.1 & 81.2 \\
        KDDG(Ours) & 81.0 & \textbf{78.7} & \textbf{96.0} & 72.3 & \textbf{82.0} \\
    \bottomrule
    \end{tabular}
    }
    \caption{Evaluation of DG on the PACS benchmark. The average target domain accuracy of five repeated experiments is reported.}
    \label{tab:pacs}
\end{table}

\begin{table*}[htbp]
  \centering
    \scalebox{1}{
    \subtable[DeepAll]{
    \begin{tabular}{c|ccccc}
    \toprule
    target & DSC  & CC    & JI & TPR   & ASD \\
    \midrule
     1     & 0.8560 & 65.34 & 0.7520 & 0.8746 & 0.0809 \\
     2     & 0.7323 & 26.21 & 0.5789 & 0.8109  & 0.0992 \\
     3     & 0.5041 & -209  & 0.3504 & 0.4926 & 1.8661\\
     4     & 0.8775 & 71.92  & 0.7827 & 0.8888 & 0.0599\\
    \midrule
    Average& 0.7425	& -11.4 & 0.6160 & 0.7667 &	0.5265
  \\
    \bottomrule
    \end{tabular}%
    }


    \subtable[CCSA \cite{yoon2019generalizable}]{
    \begin{tabular}{c|ccccc}
    \toprule
    target & DSC   & CC    & JI & TPR   & ASD \\
    \midrule
     1     & 0.8061 & 50.15 & 0.6801 & 0.8703 & 0.1678 \\
     2     & 0.8009 & 50.04 & 0.6687 & 0.8141 & 0.0939 \\
     3     & 0.5012 & -112  & 0.3389 & 0.5444 & 1.5480 \\
     4     & 0.8686 & 69.61 & 0.7684 & 0.8926 & 0.0449  \\
    \midrule
    Average& 0.7442 & 14.45 & 0.6140 & 0.7804 & 0.4637  \\
    \bottomrule
    \end{tabular}%
    }
    }
    \subtable[MASF \cite{dou2019domain}]{
    \begin{tabular}{c|ccccc}
    \toprule
    target & DSC   & CC    & JI & TPR   & ASD \\
    \midrule
     1     & 0.8502 & 64.22 & 0.7415 & 0.8903 & 0.2274 \\
     2     & 0.8115 & 53.04 & 0.6844 & 0.8161 & 0.0826 \\
     3     & 0.5285 & -99.3 & 0.3665 & 0.5155 & 1.8554 \\
     4     & \bf{0.8938} & \bf{76.14} & \bf{0.8083} & 0.8991 & 0.0366  \\
    \midrule
    Average & 0.7710 & 23.52 & 0.6502 & 0.7803 & 0.5505  \\
    \bottomrule
    \end{tabular}%
    }
    \subtable[MLDG \cite{li2018learning}]{
    \begin{tabular}{c|ccccc}
    \toprule
    target & DSC   & CC    & JI & TPR   & ASD \\
    \midrule
     1     & 0.8585 & 64.57 & 0.7489 & 0.8520 & 0.0573 \\
     2     & 0.8008 & 49.65 & 0.6696 & 0.7696 & 0.0745 \\
     3     & 0.5269 & -108  & 0.3668 & 0.5066 & 1.7708 \\
     4     & 0.8837 & 73.60 & 0.7920 & 0.8637 & 0.0451 \\
    \midrule
    Average& 0.7675 & 19.96 & 0.6443 & 0.7480 & 0.4869 \\
    \bottomrule
    \end{tabular}%
    }
    
    \subtable[\textcolor{black}{LDDG \cite{li2020domain}}]{
    \begin{tabular}{c|ccccc}
    \toprule
    target & DSC   & CC    & JI & TPR   & ASD \\
    \midrule
     1     & 0.8708 & 69.29 & 0.7753 & \bf{0.8978} & \bf{0.0411}\\
     2     & \bf{0.8364} & \bf{60.58} & \bf{0.7199} & \bf{0.8485} & \bf{0.0416} \\
     3     & 0.5543 & -71.6 & \bf{0.3889} & \bf{0.5923} & 1.5187  \\
     4     & 0.8910 & 75.46 & 0.8039 & 0.8844 & \bf{0.0289} \\
    \midrule
    Average& 0.7881 & 33.43 & \textbf{0.6720} & 0.8058 & 0.4076  \\
    \bottomrule
    \end{tabular}%
    }
    \subtable[KDDG (Ours)]{
    \begin{tabular}{c|ccccc}
    \toprule
    target & DSC   & CC    & JI & TPR   & ASD \\
    \midrule
     1     & \bf{0.8745} & \bf{70.75} & \bf{0.7795} & 0.8949 & 0.0539 \\
     2     & 0.8229 & 56.71 & 0.6997 & 0.8226 & 0.04901 \\
     3     & \bf{0.5676} & \bf{-63.1} & 0.3866 & 0.5904 & \bf{1.2805} \\
     4     & 0.8894 & 75.06 & 0.8011 & \bf{0.9222} & 0.0377  \\
    \midrule
    Average& \bf{0.7886} & \bf{34.86} & 0.6667 & \bf{0.8075} & \bf{0.3553}  \\
    \bottomrule
    \end{tabular}%
    }

  \caption{Domain generalization results on gray matter segmentation task. }
\label{tab:gm_result}%
\end{table*}

\subsection{Mini-DomainNet}

We then consider mini-DomainNet \cite{zhou2020domain}, which is a subset of DomainNet \cite{peng2019moment} for evaluation. It contains 4 domains, namely sketch, real, clipart and painting, with more than 140k images in total.

\textbf{Settings:} We use SGD with momentum as the optimizer. The initial learning rate is 5e-3 and is decreased using the cosine annealing rule \cite{loshchilov2016sgdr}. The batch size is 128 with a random sampler from the concatenated source domains. Resnet-18 is used as the backbone network for all competitors and we train the model for 60 epochs. We use the same data augmentation for all the methods, including random flip and random crop with a scale factor of 1.25. We adjusted the hyperparameters for the competitors in a wide range and report the best results we can achieve here.

\textbf{Results:} We repeat the experiments for 3 times and report the average accuracy in Table \ref{tab:minidomainnet}. We can find that our method has an improvement in a clear margin compared with other methods in the mini-DomainNet benchmark which is much larger than PACS. Such observation further justifies the significance of our proposed method to handle large-scale data.
\begin{table}[]
    \centering
    \scalebox{1}{
    \begin{tabular}{cccccc}
    \toprule
        & Clipart & Real & Painting & Sketch & Avg.  \\
    \midrule
        DeepAll & 62.86 & 58.73 & 47.94 & 43.02 & 53.14\\ 
        MLDG \cite{li2018learning}& 63.54 & 59.49 & 48.68 & 43.41 & 53.78\\
        JiGen \cite{carlucci2019domain} & 63.84 & 58.80 & \textbf{49.40} & 44.26 & 54.08\\
        MASF \cite{dou2019domain} & 63.05 & 59.22 & 48.34 & 43.67 & 53.58 \\
        RSC \cite{huang2020self} & 64.65 & 59.37 & 46.71 & 42.38 & 53.94 \\ 
        KDDG(Ours) & \textbf{65.80} & \textbf{62.06} & 49.37 & \textbf{46.19} & \textbf{55.86} \\
    \bottomrule
    \end{tabular}
    }
    \caption{Evaluation of DG on mini-DomainNet benchmark. }
    \label{tab:minidomainnet}
\end{table}

\subsection{Gray Matter Segmentation}
 We further evaluate our proposed method on gray matter segmentation task \cite{li2020domain, prados2017spinal} which aims to segment the gray matter area of the spinal cord for medical diagnosis.
We use the data from spinal cord gray matter segmentation challenge \cite{prados2017spinal} which are collected from four different medical centers with different MRI scanner parameters, e.g., different resolution from $0.25 \times 0.25 \times 2.5mm$ to $0.5 \times 0.5 \times 5mm$, the flip angle from $7^{\circ}$ to  $35^{\circ}$, different coil type. These differences lead to four different domains named 'set1', 'set2', 'set3', and 'set4'.

\textbf{Settings: } We use the same 2D-UNet \cite{ronneberger2015u, li2020domain} as the backbone for a fair comparison. A two-stage strategy in a coarse-to-fine manner is adopted following \cite{prados2017spinal, li2020domain}. More specifically, we first segment the area of the spinal cord and then extract the area of gray matter from the spinal cord. 

\begin{figure*}[htbp]
    \centering
    \includegraphics[clip, trim =0 20 0 10, width=0.8\linewidth]{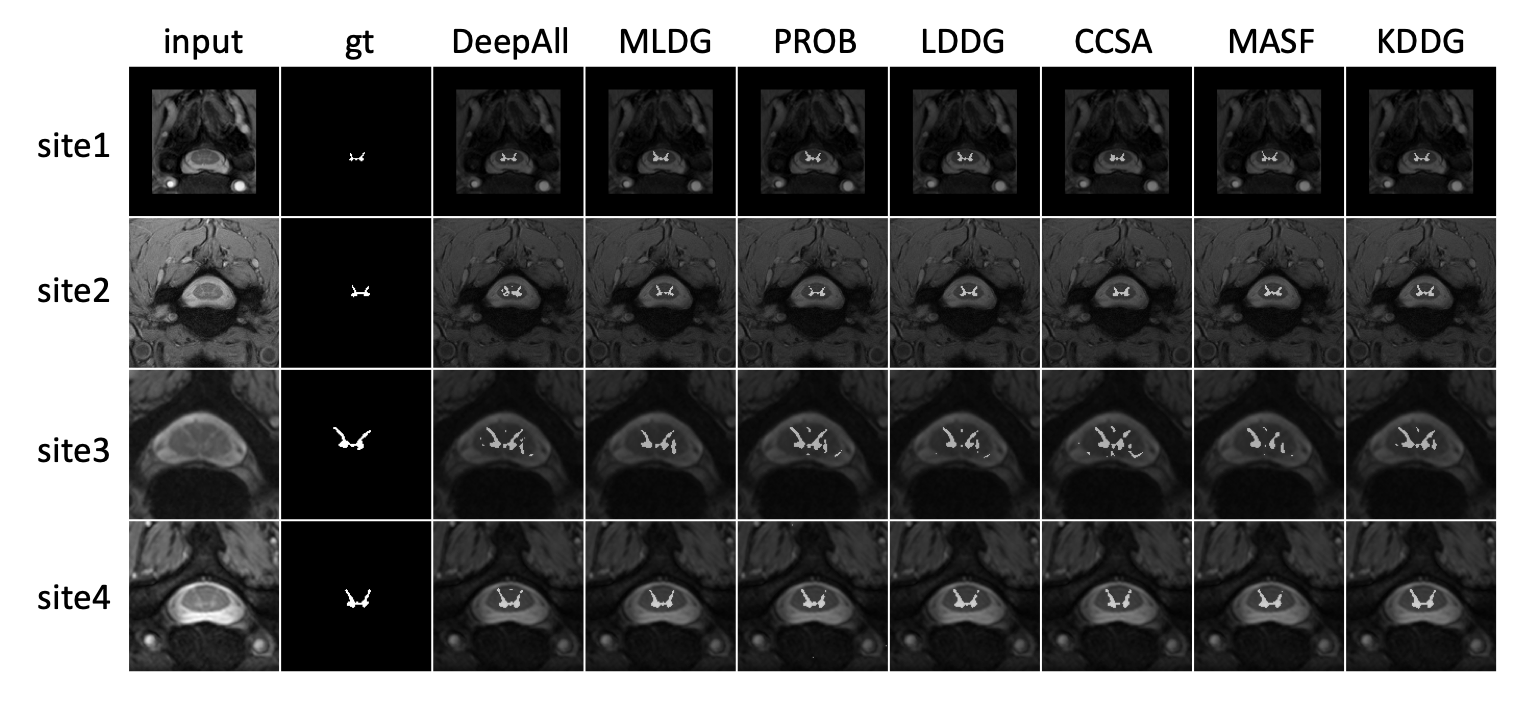}
    
    \caption{Qualitative comparisons. Each row represents a sample from a specific domain. Each column denotes the input, ground truth (gt) or different methods including DeepAll,  MLDG \cite{li2018learning}, Probabilistic U-Net \cite{kohl2018probabilistic} (abbreviated as PROB here), LDDG \cite{li2020domain}, CCSA \cite{yoon2019generalizable}, MASF \cite{dou2019domain} and KDDG, respectively.} 
    \label{fig:seg_result}
\end{figure*}

\textbf{Results:}
We compare our method with state of the art domain generalization methods, including MASF \cite{dou2019domain}, MLDG \cite{li2018learning}, CCSA \cite{yoon2019generalizable}, LDDG \cite{li2020domain}, using different metrics including three overlapping metrics: Dice Similarity Coefficient (DSC $\uparrow$), Conformity Coefficient (CC $\uparrow$), Jaccard Index (JI $\uparrow$); one statistical based metrics: True Positive Rate (TPR $\uparrow$); one distance based metric based on 3D: Average surface distance (ASD $\downarrow$). $\uparrow$ means the higher the better, $\downarrow$ means the lower the better. The results are shown in Table \ref{tab:gm_result} and Fig. \ref{fig:seg_result}. According to the results, we can find that our method has the best performance overall comparing with the latest SOTA works.

\subsection{Mountain Car}
Besides evaluating on the benchmark datasets in computer vision, we are also interested in analyzing our proposed KDDG in the reinforcement learning  task. To be specific, we use a standard reinforcement learning benchmark in OpenAI gym \cite{brockman2016openai} named mountain car. The target is to let the mountain car hit the peak with the least fuel cost with the actions ``push left", ``no push", and ``push right" in the action space.


\textbf{Settings: } We create different domains by change the gravitational constant $g$ in the game. More specifically, $g$ is set in a range from 0.0019 to 0.0031 with a step 0.0003, which leads to 5 different environments regarded as 5 domains. We use the setting of single domain generalization here in which only one domain is used for training, where $g=0.0025$, and others are used as unseen target domains for the test. We use the DQNs as the baseline and use the same backbone for all methods. The details of our backbone can be found in the supplementary materials. We compare our method with MLDG \cite{li2018learning} and  Maximum Entropy Reinforcement Learning (MERL) based method which also encourages the output to be soft and increases the generalization ability.

\textbf{Results: } We repeat the experiments for 10 times and then report the average score of fuel consumption and the corresponding standard deviation in the Table \ref{tab:mountain-car}. The smaller the value in the table, the better. It represents the amount of fuel consumption. We can find that both MLDG \cite{li2018learning} and MERL  \cite{haarnoja2018soft} have some improvement even if we use a single domain for training. We can also observe that our method has a better performance in most of the cases and has a more stable performance than the baseline method. 


\begin{table}[htbp]
\centering
\scalebox{1}{
\begin{tabular}{cccccc}
\toprule
 $g$ & Baseline & MLDG \cite{li2018learning} & MERL \cite{haarnoja2018soft} & KDDG \\
 \hline
 $0.0019$ & 111.8$\pm$1.84 & 109.9$\pm$2.43 & 110.2$\pm$1.92 & \textbf{108.7}$\pm$2.23\\ 
 $0.0022$ & 109.2$\pm$1.28 & \textbf{106.8}$\pm$1.49 & 108.1$\pm$1.97 & 107.9$\pm$1.54 \\
 $0.0025$  & 113.0$\pm$6.08 & 110.8$\pm$7.01 & 112.4$\pm$8.65 & \textbf{109.3}$\pm$4.78 \\
 $0.0028$ & 174.6$\pm$22.9 & 121.4$\pm$21.7 & 127.9$\pm$18.3 & \textbf{117.0}$\pm$20.2\\
 $0.0031$ & 167.7$\pm$11.0 & 139.8$\pm$22.1 & 134.8$\pm$24.6 & \textbf{126.8}$\pm$25.9\\
 \bottomrule
\end{tabular}
}

\caption{Single domain generalization results on the RL task moutain car. In the source domain, $g$ is set to 0.0025. The amount of fuel consumption is used as the metric \cite{brockman2016openai} for which the smaller the better.}
\label{tab:mountain-car}
\end{table}


\subsection{Ablation Study}


To further understand the contribution of each component, we conduct the ablation study by using the PACS benchmark. The results are shown in Table \ref{tab:ablation}. We can find about 1.2\% improvement from 79.5\% to 80.7\% in average when only using the gradient filter and 1.4\% improvement when only using the vanilla knowledge distillation which is competitive to the results of some DG methods. Our proposed KDDG can achieve the best performance with 2.5\% improvement in average. In addition,  we also consider \cite{muller2019does} which aims at label smoothing $y_k^{LS}=(1-\alpha)y_k+\alpha/K$ for model calibration and generalization improvement. We can find \cite{muller2019does} slightly improve the performance by 0.5\%, which is lower than using the informative label from the teacher network. Such results suggest that \cite{muller2019does} may not be helpful with cross-domain generalization improvement. 
\begin{table}[htbp]
    \centering
    \scalebox{1}{
    \begin{tabular}{cccccc}
    \toprule
        & Art & Cartoon & Photo & Sketch & Avg.  \\
    \midrule
        DeepAll & 77.0 & 75.9 & 95.5 & 70.3 & 79.5\\ 
        SoftLabel\cite{muller2019does} & 80.4 & 66.8 & 95.8 & 76.9 & 80.0 \\
        KD & 80.0 & 77.9 & 95.4 & 70.2 & 80.9 \\
        GradFilter & 78.9 & 77.8 & 95.7 & 70.5 & 80.7\\
        KDDG & \textbf{81.0} & \textbf{78.7} & \textbf{96.0} & \textbf{72.3} & \textbf{82.0} \\
    \bottomrule
    \end{tabular}
    }

    \caption{Ablation study on the PACS benchmark. KD: using vanilla knowledge distillation in Eq.\ref{student_loss}. GradFilter: only use the gradient filter without a distillation manner. KDDG: our complete version.}
    \label{tab:ablation}
\end{table}

\section{Further Analysis}
To further explore the rationale of the effectiveness of our proposed method, we propose two metrics to quantify the task difficulty and the amount of superfluous domain specific information respectively.
\subsection{Difficulty Quantification}
\subsubsection{Metric to Quantify the Difficulty}
\label{sec:quantify}
As discussed previously, there exists a connection between the task difficulty and the generalization capability of the trained model. We have shown our preliminary experiments in Fig. \ref{fig:noise-level} to explore the impact of task difficulty on generalization performance by setting different noisy label levels of the ground truth. We are also interested to quantify the task difficulty to analyze how task difficulty influences the generalization capability.  While there exist some works to quantify the difficulty in computer vision tasks \cite{tudor2016hard, nie2019difficulty}, however, they mainly focus on assessing the difficulty of a single picture or of a certain area in the picture \cite{nie2019difficulty, difficulty_sr}, which may not be suitable in our setting.  To this end, we propose to use \textbf{cumulative weight distance ($\text{CWD}$)} to quantify the difficulty of the task, where $\text{CWD}$ can be defined as

\begin{equation}
    \text{CWD}=\sum_{k=1}^{N} \left \| w_k - w_{k-1} \right \|_2
\end{equation}
where $w_k$ is the parameters expanded into a vector at epoch $k$ and $w_0$ denotes the initial parameters vector. $N$ denotes the maximum number of epochs. We adopt the metric $\text{CWD}$ to verify our assumption that the simpler the task is, the smaller $\text{CWD}$ is. 

\subsubsection{Analysis of Results Based on the Cumulative Weight Distance}

\begin{figure}
    \centering
    \scalebox{0.7}{
%
%
\definecolor{mycolor1}{rgb}{0.00000,0.44700,0.74100}%
\definecolor{mycolor2}{rgb}{0.85000,0.32500,0.09800}%
\definecolor{mycolor3}{rgb}{0.92900,0.69400,0.12500}%
\definecolor{mycolor4}{rgb}{0.49400,0.18400,0.55600}%
\begin{tikzpicture}

\begin{axis}[%
width=2.9in,
height=1.7in,
at={(0in,0in)},
scale only axis,
xmin=0,
xmax=50,
xlabel style={font=\color{white!15!black}},
xlabel={epoch},
ymin=0,
ymax=30,
ylabel style={font=\color{white!15!black}},
ylabel={CWD},
axis background/.style={fill=white},
legend style={at={(0.97,0.03)}, anchor=south east, legend cell align=left, align=left, draw=white!15!black}
]
\addplot [color=mycolor1, dashed, line width=2.0pt]
  table[row sep=crcr]{%
1	2.631982803\\
2	4.348534107\\
3	5.710112572\\
4	6.88879776\\
5	7.919690609\\
6	8.863217354\\
7	9.750803947\\
8	10.56381893\\
9	11.35269737\\
10	12.12061787\\
11	12.86144733\\
12	13.60775757\\
13	14.3451376\\
14	15.0768137\\
15	15.82973385\\
16	16.55895996\\
17	17.29769325\\
18	18.04406357\\
19	18.74061775\\
20	19.3962326\\
21	20.00591087\\
22	20.58405685\\
23	21.11495209\\
24	21.64055824\\
25	22.11855888\\
26	22.53966522\\
27	22.92890358\\
28	23.27836609\\
29	23.62197304\\
30	23.91903496\\
31	24.17361832\\
32	24.39581108\\
33	24.59784126\\
34	24.76308632\\
35	24.90456581\\
36	25.01498413\\
37	25.11318207\\
38	25.20031357\\
39	25.27249718\\
40	25.33168983\\
41	25.38065147\\
42	25.41769981\\
43	25.44609261\\
44	25.46908188\\
45	25.48703957\\
46	25.49852371\\
47	25.50578499\\
48	25.51012802\\
49	25.51219749\\
50	25.51270485\\
};
\addlegendentry{$\lambda\text{=0.3}$}

\addplot [color=mycolor2, dotted, line width=2.0pt]
  table[row sep=crcr]{%
1	2.649866104\\
2	4.557481766\\
3	5.945204258\\
4	7.144420624\\
5	8.182362556\\
6	9.170254707\\
7	10.05925179\\
8	10.92934704\\
9	11.78290081\\
10	12.58908081\\
11	13.36671829\\
12	14.15204525\\
13	14.92037106\\
14	15.66980934\\
15	16.44429779\\
16	17.16637993\\
17	17.8268261\\
18	18.50511169\\
19	19.11617279\\
20	19.69693756\\
21	20.2444725\\
22	20.73346901\\
23	21.23112488\\
24	21.65946198\\
25	22.06773758\\
26	22.43435097\\
27	22.75474548\\
28	23.03364182\\
29	23.29457283\\
30	23.50611496\\
31	23.70207405\\
32	23.86526489\\
33	24.00061607\\
34	24.1240406\\
35	24.23135185\\
36	24.32154846\\
37	24.40850449\\
38	24.47727203\\
39	24.53771591\\
40	24.589468\\
41	24.62952232\\
42	24.66306686\\
43	24.69147491\\
44	24.71195412\\
45	24.72767258\\
46	24.73895836\\
47	24.74619865\\
48	24.7502327\\
49	24.75195503\\
50	24.75238419\\
};
\addlegendentry{$\lambda\text{=0.2}$}

\addplot [color=mycolor3, dashdotted, line width=2.0pt]
  table[row sep=crcr]{%
1	2.660161734\\
2	4.586120129\\
3	6.285271645\\
4	7.73585844\\
5	9.009652138\\
6	10.14232445\\
7	11.17445469\\
8	12.1364603\\
9	13.03810406\\
10	13.89706135\\
11	14.69981575\\
12	15.46612453\\
13	16.1706562\\
14	16.86169815\\
15	17.48057175\\
16	18.08633041\\
17	18.67185402\\
18	19.20432663\\
19	19.71697235\\
20	20.17656708\\
21	20.58456039\\
22	20.99174118\\
23	21.34090614\\
24	21.67695808\\
25	21.96906853\\
26	22.22883415\\
27	22.45175934\\
28	22.6515007\\
29	22.83096504\\
30	22.97579956\\
31	23.10835648\\
32	23.23234558\\
33	23.33503342\\
34	23.43151665\\
35	23.51359177\\
36	23.58955956\\
37	23.65693474\\
38	23.71622658\\
39	23.76482391\\
40	23.8073616\\
41	23.84832764\\
42	23.87894249\\
43	23.90184784\\
44	23.91917229\\
45	23.93189621\\
46	23.94067192\\
47	23.94626617\\
48	23.94922066\\
49	23.95056915\\
50	23.95092583\\
};
\addlegendentry{$\lambda\text{=0.1}$}

\addplot [color=mycolor4, line width=2.0pt]
  table[row sep=crcr]{%
1	2.807541609\\
2	4.11251688\\
3	5.166558266\\
4	6.061757565\\
5	6.862466812\\
6	7.57574892\\
7	8.236385345\\
8	8.820618629\\
9	9.354649544\\
10	9.835541725\\
11	10.29904461\\
12	10.72027302\\
13	11.10201263\\
14	11.42577744\\
15	11.72169018\\
16	12.00468922\\
17	12.25827122\\
18	12.48888683\\
19	12.69214916\\
20	12.83419323\\
21	12.99195957\\
22	13.1287899\\
23	13.26211357\\
24	13.38065052\\
25	13.49925518\\
26	13.60381985\\
27	13.69970322\\
28	13.78679466\\
29	13.87024403\\
30	13.94866562\\
31	14.01685238\\
32	14.08335781\\
33	14.14192963\\
34	14.1917429\\
35	14.24185753\\
36	14.28497028\\
37	14.3230648\\
38	14.35692787\\
39	14.38746834\\
40	14.41305351\\
41	14.43621159\\
42	14.45546818\\
43	14.47041321\\
44	14.48316097\\
45	14.4926157\\
46	14.4991293\\
47	14.50332165\\
48	14.50612831\\
49	14.50761414\\
50	14.50796223\\
};
\addlegendentry{$\lambda\text{=0}$}

\end{axis}

\begin{axis}[%
width=2.9in,
height=1.7in,
at={(0in,0in)},
scale only axis,
xmin=0,
xmax=1,
ymin=0,
ymax=1,
axis line style={draw=none},
ticks=none,
axis x line*=bottom,
axis y line*=left
]
\end{axis}
\end{tikzpicture}

    }
    \caption{The change of CWD with training using different noise level $\lambda$. We use syn in digit-five as the target domain and train the model for 50 epochs. We can find that CWD and task difficulty are positively correlated, i.e., the easier the task, the smaller CWD.}
    \label{fig:cwd}
\end{figure}
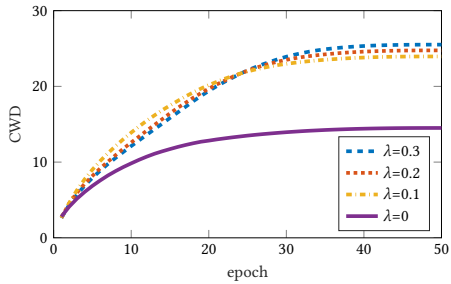

The results of CWD by varying noise level $\lambda$ are shown in Fig. \ref{fig:cwd} using the same setting as Fig. \ref{fig:insight}. We can draw the conclusion that the simpler the task, the smaller $\text{CWD}$ and the better generalization. To further explore the impact of our method KDDG on CWD, we also conduct experiments using different components of our method. The results are shown in Table \ref{tab:cwd}. We can find that KD, GradFilter, and KDDG can reduce the CWD compared with directly training with the DeepAll baseline. Such observation further verifies our assumption. It’s worth noting that there is a lower bound of CWD to learn the domain invariant feature still, so here we emphasize the difference of CWD between baseline and our methods. 
\begin{table}[htbp]
    \centering
    \scalebox{1}{
    \begin{tabular}{cccccc}
    \toprule
        & DeepAll & KD & GradFilter & KDDG  \\
    \midrule
    CWD & 14.51 & 14.44  & 14.30  & 14.28 \\
    \bottomrule
    \end{tabular}
    }
    \caption{The impact of different components on CWD.}
    \label{tab:cwd}
\end{table}

\subsection{Quantification of Domain Specific Information}
\subsubsection{Metric Definition}
We further analyze how our proposed method benefits the generalization capability of the model. Specifically, we use mutual information $I(Z,Y_d)$ to quantify the amount of residual domain specific information where $Z$ is the latent feature space and $Y_d$ is the domain label space. The idea is that, the less domain-specific information contained in the feature space, the more shareable information we can obtain which benefit domain generalization.

Mutual information between $Z$ and $Y_d$ is defined as
\begin{equation}
    I(Z,Y_d) = \int dy_d \, dz \, f(y_d,z) \log \frac{ f(y_d,z)}{f(y_d) f(z)}.
\end{equation}
We propose to use the Monte Carlo method such that the mutual information $I(Z,Y_d)$ in a tractable manner. Specifically, $I(Z,Y_d)$ can be derived as

\begin{eqnarray}
\begin{aligned}
    I(Z,Y_d) &= \int dy_d \, dz \, f(y_d,z) \log \frac{ f(y_d,z)}{f(y_d) f(z)} \\
           &=\!\int \!\!dy_d \, dz \, f(y_d, z) \log f(y_d|z) \! \\
           & -\!\! \int \!dy_d \, f(y_d) \log f(y_d) \\
           &= H(Y_d)-\mathbb{E}_{z\sim f(z)}H(f(y_d|z)) .
\end{aligned}
\end{eqnarray}


To compute $I(Z,Y_d)$, we also need to compute $f(z)$, which is the distribution of the latent feature $z$, as well as the likelihood of the domain discriminator $f(y|z)$. $f(z)$ can be computed by using the law of total probability \cite{zwillinger1999crc}
\begin{equation}
    f(z)=\sum_{i=1}^{N} f(z|x_i) p(x_i),
\end{equation}
where $f(z|x)$ is the conditional distribution from the feature encoding sub-network of the original model, $f(y_d|z)$ can be computed by using SVM with class probability \cite{wu2004probability} with five cross validation in the source domains. We empirically find that it has faster speed, higher accuracy and better stability than convolutional neural networks based domain discriminators.

\subsubsection{Analysis of Results Based on Mutual Information}
\begin{figure}[htbp]
\begin{center}
\scalebox{0.7}{
    \centering
%
%
\definecolor{mycolor1}{rgb}{0.00000,0.44706,0.74118}%
\definecolor{mycolor2}{rgb}{0.85000,0.32500,0.09800}%
\definecolor{mycolor3}{rgb}{0.92900,0.69400,0.12500}%
\definecolor{mycolor4}{rgb}{0.49400,0.18400,0.55600}%
\begin{tikzpicture}

\begin{axis}[%
width=3.3in,
height=1.8in,
at={(-0.35in,0in)},
scale only axis,
xmin=0,
xmax=27,
xlabel style={font=\color{white!15!black}},
xlabel={epoch},
ymin=0.7,
ymax=1.3,
ylabel style={font=\color{white!15!black}},
ylabel={$\text{I(Z, Y}_\text{d}\text{)}$},
axis background/.style={fill=white},
axis x line*=bottom,
axis y line*=left,
legend style={legend cell align=left, align=left, draw=white!15!black},
]
\addplot [color=mycolor1, dashdotted, line width=2.0pt]
  table[row sep=crcr]{%
1	1.12033623179967\\
2	1.08890608377182\\
3	1.06459135377311\\
4	1.04792270734636\\
5	1.02756874505047\\
6	1.00591662743506\\
7	0.982327559808268\\
8	0.953498345170753\\
9	0.927159501552108\\
10	0.921960655313521\\
11	0.934629461702278\\
12	0.9488378300631\\
13	0.94964087518109\\
14	0.935550537597608\\
15	0.912127817485347\\
16	0.893591634839279\\
17	0.889378522253297\\
18	0.898123343893208\\
19	0.906253400503643\\
20	0.906376598269195\\
21	0.902489704633456\\
22	0.902124297159327\\
23	0.905270656757508\\
24	0.908656903190524\\
25	0.910639870740191\\
26	0.911737633794984\\
};
\addlegendentry{$\lambda\text{ = 0 (DeepAll)}$}

\addplot [color=mycolor2, dotted, line width=2.0pt]
  table[row sep=crcr]{%
1	1.12612264007569\\
2	1.1056236088337\\
3	1.07788464976405\\
4	1.05062434782603\\
5	1.02554193265163\\
6	1.00726932051145\\
7	1.00217453251845\\
8	1.01602608343737\\
9	1.03109022787726\\
10	1.04358190783493\\
11	1.04460871887943\\
12	1.02331688587733\\
13	0.986194368781263\\
14	0.957744618002548\\
15	0.945009659249721\\
16	0.938392094676507\\
17	0.922158796106955\\
18	0.898507254497823\\
19	0.887891889588049\\
20	0.899523977484255\\
21	0.921015753117602\\
22	0.935114043188437\\
23	0.940539440455674\\
24	0.942883171096434\\
25	0.952862039800736\\
26	0.968223663851708\\
};
\addlegendentry{$\lambda\text{ = 0.15 (DeepAll)}$}

\addplot [color=mycolor3, dashed, line width=2.0pt]
  table[row sep=crcr]{%
1	1.11107390618482\\
2	1.0908540213897\\
3	1.07173307539444\\
4	1.06003658842607\\
5	1.05997149247714\\
6	1.06362839285357\\
7	1.05592705986205\\
8	1.04412073008113\\
9	1.0375302071032\\
10	1.03091933911387\\
11	1.0198825631184\\
12	1.01062788793524\\
13	1.0084738585015\\
14	1.01634884493499\\
15	1.02716063117596\\
16	1.01714765183687\\
17	0.981843335714625\\
18	0.945751553959141\\
19	0.930590743672036\\
20	0.939707752066108\\
21	0.954483661940768\\
22	0.963576089647267\\
23	0.971543142358407\\
24	0.976978035288119\\
25	0.983118585282828\\
26	0.99212260502366\\
};
\addlegendentry{$\lambda\text{ = 0.30 (DeepAll)}$}

\addplot [color=mycolor4, line width=2.0pt]
  table[row sep=crcr]{%
1	1.04807196056777\\
2	1.01571186193612\\
3	0.9895328281974\\
4	0.961982841848784\\
5	0.921877913363756\\
6	0.878294280733393\\
7	0.845004174940404\\
8	0.824225368413457\\
9	0.805298176472849\\
10	0.785887142837721\\
11	0.772766863471678\\
12	0.771679409999098\\
13	0.79008319415643\\
14	0.81210562811234\\
15	0.812697070885602\\
16	0.787509871140914\\
17	0.759871861468116\\
18	0.745322157750064\\
19	0.742624876855311\\
20	0.745685644538536\\
21	0.747159729714206\\
22	0.747602348177188\\
23	0.751847190337447\\
24	0.750943157705315\\
25	0.742462913648752\\
26	0.733196690930612\\
};
\addlegendentry{$\lambda\text{ = 0 (KDDG)}$}

\end{axis}
\end{tikzpicture}%
}
\end{center}
\caption{Trends in the amount of residual domain specific information during training under different noise level $\lambda$. Our method KDDG significantly reduce the domain specific information comparing with DeepAll.}
\label{fig:mutal_info}
\end{figure}
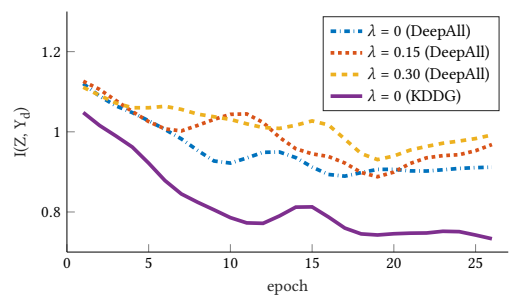

Now we use the metric $I(Z,Y_d)$ to quantify the amount of domain-specific information remaining in the network. The sketch domain in the PACS benchmark is used for evaluation on account of its good discrimination. We train the models for 26 epochs and show the amount of domain-specific information $I(Z,Y_d)$ remaining in the network in Fig. \ref{fig:mutal_info}. As we can see, domain specific information will quickly decrease at the beginning of the training phase. However, as the training progresses, the model will inevitably overfit on the training source domains. Besides, we can also draw a conclusion from Fig. \ref{fig:mutal_info} that the harder the task, the more domain specific information will be learned. Last, we show the results of $I(Z,Y_d)$ in Table \ref{tab:mutual_info} by considering different component. We can see that by jointly considering KD and GradFilter, more domain-specific information can be removed. 

\begin{table}[tbp]
    \centering
    \scalebox{1}{
    \begin{tabular}{cccccc}
    \toprule
        & DeepAll & KD & GradFilter & KDDG  \\
    \midrule
    $I(Z,Y_d)$ & 0.91 & 0.82 & 0.86 & 0.73 \\
    \bottomrule
    \end{tabular}
    }
    \label{tab:mutual_info}
    \caption{The amount of domain-specific information remaining in the network. KD: using vanilla knowledge distillation in Eq.\ref{student_loss} from vanilla teacher.} 
    \label{tab:mutual_info}
\end{table}

\section{Conclusion}
We address the domain generalization problem by proposing a simple, effective, and plug-and-play training strategy based on a knowledge distillation framework with a novel gradient regularization. We show that our method can achieve state-of-the-art performance on five different DG benchmarks including classification, segmentation, and reinforcement learning. We further analyze the significance and effectiveness of our method by proposing two metrics from the perspective of task difficulty and domain information.

\bibliographystyle{ACM-Reference-Format}
\bibliography{acmart.bbl}


\begin{thebibliography}{66}


\ifx \showCODEN    \undefined \def \showCODEN     #1{\unskip}     \fi
\ifx \showDOI      \undefined \def \showDOI       #1{#1}\fi
\ifx \showISBNx    \undefined \def \showISBNx     #1{\unskip}     \fi
\ifx \showISBNxiii \undefined \def \showISBNxiii  #1{\unskip}     \fi
\ifx \showISSN     \undefined \def \showISSN      #1{\unskip}     \fi
\ifx \showLCCN     \undefined \def \showLCCN      #1{\unskip}     \fi
\ifx \shownote     \undefined \def \shownote      #1{#1}          \fi
\ifx \showarticletitle \undefined \def \showarticletitle #1{#1}   \fi
\ifx \showURL      \undefined \def \showURL       {\relax}        \fi
\providecommand\bibfield[2]{#2}
\providecommand\bibinfo[2]{#2}
\providecommand\natexlab[1]{#1}
\providecommand\showeprint[2][]{arXiv:#2}

\bibitem[\protect\citeauthoryear{Ahuja, Shanmugam, Varshney, and
  Dhurandhar}{Ahuja et~al\mbox{.}}{2020}]%
        {ahuja2020invariant}
\bibfield{author}{\bibinfo{person}{Kartik Ahuja}, \bibinfo{person}{Karthikeyan
  Shanmugam}, \bibinfo{person}{Kush Varshney}, {and} \bibinfo{person}{Amit
  Dhurandhar}.} \bibinfo{year}{2020}\natexlab{}.
\newblock \showarticletitle{Invariant risk minimization games}.
\newblock \bibinfo{journal}{\emph{arXiv preprint arXiv:2002.04692}}
  (\bibinfo{year}{2020}).
\newblock


\bibitem[\protect\citeauthoryear{Arjovsky, Bottou, Gulrajani, and
  Lopez-Paz}{Arjovsky et~al\mbox{.}}{2019}]%
        {arjovsky2019invariant}
\bibfield{author}{\bibinfo{person}{Martin Arjovsky}, \bibinfo{person}{L{\'e}on
  Bottou}, \bibinfo{person}{Ishaan Gulrajani}, {and} \bibinfo{person}{David
  Lopez-Paz}.} \bibinfo{year}{2019}\natexlab{}.
\newblock \showarticletitle{Invariant risk minimization}.
\newblock \bibinfo{journal}{\emph{arXiv preprint arXiv:1907.02893}}
  (\bibinfo{year}{2019}).
\newblock


\bibitem[\protect\citeauthoryear{Balaji, Sankaranarayanan, and
  Chellappa}{Balaji et~al\mbox{.}}{2018}]%
        {balaji2018metareg}
\bibfield{author}{\bibinfo{person}{Yogesh Balaji}, \bibinfo{person}{Swami
  Sankaranarayanan}, {and} \bibinfo{person}{Rama Chellappa}.}
  \bibinfo{year}{2018}\natexlab{}.
\newblock \showarticletitle{MetaReg: Towards Domain Generalization using
  Meta-Regularization}. In \bibinfo{booktitle}{\emph{NuerIPS}}.
  \bibinfo{pages}{998--1008}.
\newblock


\bibitem[\protect\citeauthoryear{Berthelot, Raffel, Roy, and
  Goodfellow}{Berthelot et~al\mbox{.}}{2019}]%
        {berthelot2018understanding}
\bibfield{author}{\bibinfo{person}{David Berthelot}, \bibinfo{person}{Colin
  Raffel}, \bibinfo{person}{Aurko Roy}, {and} \bibinfo{person}{Ian
  Goodfellow}.} \bibinfo{year}{2019}\natexlab{}.
\newblock \showarticletitle{Understanding and improving interpolation in
  autoencoders via an adversarial regularizer}.
\newblock \bibinfo{journal}{\emph{ICLR}} (\bibinfo{year}{2019}).
\newblock


\bibitem[\protect\citeauthoryear{Bousmalis, Silberman, Dohan, Erhan, and
  Krishnan}{Bousmalis et~al\mbox{.}}{2017}]%
        {bousmalis2017unsupervised}
\bibfield{author}{\bibinfo{person}{Konstantinos Bousmalis},
  \bibinfo{person}{Nathan Silberman}, \bibinfo{person}{David Dohan},
  \bibinfo{person}{Dumitru Erhan}, {and} \bibinfo{person}{Dilip Krishnan}.}
  \bibinfo{year}{2017}\natexlab{}.
\newblock \showarticletitle{Unsupervised pixel-level domain adaptation with
  generative adversarial networks}. In \bibinfo{booktitle}{\emph{CVPR}},
  Vol.~\bibinfo{volume}{1}. \bibinfo{pages}{7}.
\newblock


\bibitem[\protect\citeauthoryear{Brockman, Cheung, Pettersson, Schneider,
  Schulman, Tang, and Zaremba}{Brockman et~al\mbox{.}}{2016}]%
        {brockman2016openai}
\bibfield{author}{\bibinfo{person}{Greg Brockman}, \bibinfo{person}{Vicki
  Cheung}, \bibinfo{person}{Ludwig Pettersson}, \bibinfo{person}{Jonas
  Schneider}, \bibinfo{person}{John Schulman}, \bibinfo{person}{Jie Tang},
  {and} \bibinfo{person}{Wojciech Zaremba}.} \bibinfo{year}{2016}\natexlab{}.
\newblock \showarticletitle{Openai gym}.
\newblock \bibinfo{journal}{\emph{arXiv preprint arXiv:1606.01540}}
  (\bibinfo{year}{2016}).
\newblock


\bibitem[\protect\citeauthoryear{Buciluǎ, Caruana, and
  Niculescu-Mizil}{Buciluǎ et~al\mbox{.}}{2006}]%
        {bucilu2006model}
\bibfield{author}{\bibinfo{person}{Cristian Buciluǎ}, \bibinfo{person}{Rich
  Caruana}, {and} \bibinfo{person}{Alexandru Niculescu-Mizil}.}
  \bibinfo{year}{2006}\natexlab{}.
\newblock \showarticletitle{Model compression}. In
  \bibinfo{booktitle}{\emph{SIGKDD}}.
\newblock


\bibitem[\protect\citeauthoryear{Carlucci, D'Innocente, Bucci, Caputo, and
  Tommasi}{Carlucci et~al\mbox{.}}{2019}]%
        {carlucci2019domain}
\bibfield{author}{\bibinfo{person}{Fabio~Maria Carlucci},
  \bibinfo{person}{Antonio D'Innocente}, \bibinfo{person}{Silvia Bucci},
  \bibinfo{person}{Barbara Caputo}, {and} \bibinfo{person}{Tatiana Tommasi}.}
  \bibinfo{year}{2019}\natexlab{}.
\newblock \showarticletitle{Domain Generalization by Solving Jigsaw Puzzles}.
\newblock \bibinfo{journal}{\emph{arXiv preprint arXiv:1903.06864}}
  (\bibinfo{year}{2019}).
\newblock


\bibitem[\protect\citeauthoryear{Cheng, Rao, Chen, and Zhang}{Cheng
  et~al\mbox{.}}{2020}]%
        {cheng2020explaining}
\bibfield{author}{\bibinfo{person}{Xu Cheng}, \bibinfo{person}{Zhefan Rao},
  \bibinfo{person}{Yilan Chen}, {and} \bibinfo{person}{Quanshi Zhang}.}
  \bibinfo{year}{2020}\natexlab{}.
\newblock \showarticletitle{Explaining Knowledge Distillation by Quantifying
  the Knowledge}. In \bibinfo{booktitle}{\emph{CVPR}}.
\newblock


\bibitem[\protect\citeauthoryear{Dou, de~Castro, Kamnitsas, and Glocker}{Dou
  et~al\mbox{.}}{2019}]%
        {dou2019domain}
\bibfield{author}{\bibinfo{person}{Qi Dou}, \bibinfo{person}{Daniel~Coelho de
  Castro}, \bibinfo{person}{Konstantinos Kamnitsas}, {and} \bibinfo{person}{Ben
  Glocker}.} \bibinfo{year}{2019}\natexlab{}.
\newblock \showarticletitle{Domain generalization via model-agnostic learning
  of semantic features}. In \bibinfo{booktitle}{\emph{NuerIPS}}.
\newblock


\bibitem[\protect\citeauthoryear{Ganin, Ustinova, Ajakan, Germain, Larochelle,
  Laviolette, Marchand, and Lempitsky}{Ganin et~al\mbox{.}}{2016}]%
        {ganin2016domain}
\bibfield{author}{\bibinfo{person}{Yaroslav Ganin}, \bibinfo{person}{Evgeniya
  Ustinova}, \bibinfo{person}{Hana Ajakan}, \bibinfo{person}{Pascal Germain},
  \bibinfo{person}{Hugo Larochelle}, \bibinfo{person}{Fran{\c{c}}ois
  Laviolette}, \bibinfo{person}{Mario Marchand}, {and} \bibinfo{person}{Victor
  Lempitsky}.} \bibinfo{year}{2016}\natexlab{}.
\newblock \showarticletitle{Domain-adversarial training of neural networks}.
\newblock \bibinfo{journal}{\emph{Journal of Machine Learning Research}}
  \bibinfo{volume}{17}, \bibinfo{number}{59} (\bibinfo{year}{2016}),
  \bibinfo{pages}{1--35}.
\newblock


\bibitem[\protect\citeauthoryear{Ghifary, Balduzzi, Kleijn, and Zhang}{Ghifary
  et~al\mbox{.}}{2017}]%
        {ghifary2017scatter}
\bibfield{author}{\bibinfo{person}{Muhammad Ghifary}, \bibinfo{person}{David
  Balduzzi}, \bibinfo{person}{W~Bastiaan Kleijn}, {and}
  \bibinfo{person}{Mengjie Zhang}.} \bibinfo{year}{2017}\natexlab{}.
\newblock \showarticletitle{Scatter component analysis: A unified framework for
  domain adaptation and domain generalization}.
\newblock \bibinfo{journal}{\emph{T-PAMI}} \bibinfo{number}{1}
  (\bibinfo{year}{2017}), \bibinfo{pages}{1--1}.
\newblock


\bibitem[\protect\citeauthoryear{Ghifary, Bastiaan~Kleijn, Zhang, and
  Balduzzi}{Ghifary et~al\mbox{.}}{2015}]%
        {ghifary2015domain}
\bibfield{author}{\bibinfo{person}{Muhammad Ghifary}, \bibinfo{person}{W
  Bastiaan~Kleijn}, \bibinfo{person}{Mengjie Zhang}, {and}
  \bibinfo{person}{David Balduzzi}.} \bibinfo{year}{2015}\natexlab{}.
\newblock \showarticletitle{Domain generalization for object recognition with
  multi-task autoencoders}. In \bibinfo{booktitle}{\emph{CVPR}}.
  \bibinfo{pages}{2551--2559}.
\newblock


\bibitem[\protect\citeauthoryear{Guo, Pleiss, Sun, and Weinberger}{Guo
  et~al\mbox{.}}{2017}]%
        {guo2017calibration}
\bibfield{author}{\bibinfo{person}{Chuan Guo}, \bibinfo{person}{Geoff Pleiss},
  \bibinfo{person}{Yu Sun}, {and} \bibinfo{person}{Kilian~Q Weinberger}.}
  \bibinfo{year}{2017}\natexlab{}.
\newblock \showarticletitle{On calibration of modern neural networks}.
\newblock \bibinfo{journal}{\emph{arXiv preprint arXiv:1706.04599}}
  (\bibinfo{year}{2017}).
\newblock


\bibitem[\protect\citeauthoryear{Haarnoja, Zhou, Abbeel, and Levine}{Haarnoja
  et~al\mbox{.}}{2018}]%
        {haarnoja2018soft}
\bibfield{author}{\bibinfo{person}{Tuomas Haarnoja}, \bibinfo{person}{Aurick
  Zhou}, \bibinfo{person}{Pieter Abbeel}, {and} \bibinfo{person}{Sergey
  Levine}.} \bibinfo{year}{2018}\natexlab{}.
\newblock \showarticletitle{Soft actor-critic: Off-policy maximum entropy deep
  reinforcement learning with a stochastic actor}.
\newblock \bibinfo{journal}{\emph{arXiv preprint arXiv:1801.01290}}
  (\bibinfo{year}{2018}).
\newblock


\bibitem[\protect\citeauthoryear{Hinton, Vinyals, and Dean}{Hinton
  et~al\mbox{.}}{2015}]%
        {hinton2015distilling}
\bibfield{author}{\bibinfo{person}{Geoffrey Hinton}, \bibinfo{person}{Oriol
  Vinyals}, {and} \bibinfo{person}{Jeff Dean}.}
  \bibinfo{year}{2015}\natexlab{}.
\newblock \showarticletitle{Distilling the knowledge in a neural network}.
\newblock \bibinfo{journal}{\emph{arXiv preprint arXiv:1503.02531}}
  (\bibinfo{year}{2015}).
\newblock


\bibitem[\protect\citeauthoryear{Hoffman, Tzeng, Park, Zhu, Isola, Saenko,
  Efros, and Darrell}{Hoffman et~al\mbox{.}}{2018}]%
        {hoffman2017cycada}
\bibfield{author}{\bibinfo{person}{Judy Hoffman}, \bibinfo{person}{Eric Tzeng},
  \bibinfo{person}{Taesung Park}, \bibinfo{person}{Jun-Yan Zhu},
  \bibinfo{person}{Phillip Isola}, \bibinfo{person}{Kate Saenko},
  \bibinfo{person}{Alexei~A Efros}, {and} \bibinfo{person}{Trevor Darrell}.}
  \bibinfo{year}{2018}\natexlab{}.
\newblock \showarticletitle{Cycada: Cycle-consistent adversarial domain
  adaptation}.
\newblock \bibinfo{journal}{\emph{ICML}} (\bibinfo{year}{2018}).
\newblock


\bibitem[\protect\citeauthoryear{Huang, Gretton, Borgwardt, Sch{\"o}lkopf, and
  Smola}{Huang et~al\mbox{.}}{2006}]%
        {huang2006correcting}
\bibfield{author}{\bibinfo{person}{Jiayuan Huang}, \bibinfo{person}{Arthur
  Gretton}, \bibinfo{person}{Karsten~M Borgwardt}, \bibinfo{person}{Bernhard
  Sch{\"o}lkopf}, {and} \bibinfo{person}{Alex~J Smola}.}
  \bibinfo{year}{2006}\natexlab{}.
\newblock \showarticletitle{Correcting sample selection bias by unlabeled
  data}. In \bibinfo{booktitle}{\emph{NuerIPS}}.
\newblock


\bibitem[\protect\citeauthoryear{Huang, Wang, Xing, and Huang}{Huang
  et~al\mbox{.}}{2020}]%
        {huang2020self}
\bibfield{author}{\bibinfo{person}{Zeyi Huang}, \bibinfo{person}{Haohan Wang},
  \bibinfo{person}{Eric~P Xing}, {and} \bibinfo{person}{Dong Huang}.}
  \bibinfo{year}{2020}\natexlab{}.
\newblock \showarticletitle{Self-challenging improves cross-domain
  generalization}.
\newblock \bibinfo{journal}{\emph{arXiv preprint arXiv:2007.02454}}
  (\bibinfo{year}{2020}).
\newblock


\bibitem[\protect\citeauthoryear{Kawaguchi, Kaelbling, and Bengio}{Kawaguchi
  et~al\mbox{.}}{2017}]%
        {kawaguchi2017generalization}
\bibfield{author}{\bibinfo{person}{Kenji Kawaguchi},
  \bibinfo{person}{Leslie~Pack Kaelbling}, {and} \bibinfo{person}{Yoshua
  Bengio}.} \bibinfo{year}{2017}\natexlab{}.
\newblock \showarticletitle{Generalization in deep learning}.
\newblock \bibinfo{journal}{\emph{arXiv preprint arXiv:1710.05468}}
  (\bibinfo{year}{2017}).
\newblock


\bibitem[\protect\citeauthoryear{Kohl, Romera-Paredes, Meyer, De~Fauw, Ledsam,
  Maier-Hein, Eslami, Rezende, and Ronneberger}{Kohl et~al\mbox{.}}{2018}]%
        {kohl2018probabilistic}
\bibfield{author}{\bibinfo{person}{Simon Kohl}, \bibinfo{person}{Bernardino
  Romera-Paredes}, \bibinfo{person}{Clemens Meyer}, \bibinfo{person}{Jeffrey
  De~Fauw}, \bibinfo{person}{Joseph~R Ledsam}, \bibinfo{person}{Klaus
  Maier-Hein}, \bibinfo{person}{SM~Ali Eslami}, \bibinfo{person}{Danilo~Jimenez
  Rezende}, {and} \bibinfo{person}{Olaf Ronneberger}.}
  \bibinfo{year}{2018}\natexlab{}.
\newblock \showarticletitle{A probabilistic u-net for segmentation of ambiguous
  images}. In \bibinfo{booktitle}{\emph{NuerIPS}}.
\newblock


\bibitem[\protect\citeauthoryear{Li, Yang, Song, and Hospedales}{Li
  et~al\mbox{.}}{2017}]%
        {li2017deeper}
\bibfield{author}{\bibinfo{person}{Da Li}, \bibinfo{person}{Yongxin Yang},
  \bibinfo{person}{Yi-Zhe Song}, {and} \bibinfo{person}{Timothy~M Hospedales}.}
  \bibinfo{year}{2017}\natexlab{}.
\newblock \showarticletitle{Deeper, broader and artier domain generalization}.
  In \bibinfo{booktitle}{\emph{Proceedings of the IEEE International Conference
  on Computer Vision}}. \bibinfo{pages}{5542--5550}.
\newblock


\bibitem[\protect\citeauthoryear{Li, Yang, Song, and Hospedales}{Li
  et~al\mbox{.}}{2018b}]%
        {li2018learning}
\bibfield{author}{\bibinfo{person}{Da Li}, \bibinfo{person}{Yongxin Yang},
  \bibinfo{person}{Yi-Zhe Song}, {and} \bibinfo{person}{Timothy~M Hospedales}.}
  \bibinfo{year}{2018}\natexlab{b}.
\newblock \showarticletitle{Learning to generalize: Meta-learning for domain
  generalization}. In \bibinfo{booktitle}{\emph{AAAI}}.
\newblock


\bibitem[\protect\citeauthoryear{Li, Zhang, Yang, Liu, Song, and Hospedales}{Li
  et~al\mbox{.}}{2019}]%
        {li2019episodic}
\bibfield{author}{\bibinfo{person}{Da Li}, \bibinfo{person}{Jianshu Zhang},
  \bibinfo{person}{Yongxin Yang}, \bibinfo{person}{Cong Liu},
  \bibinfo{person}{Yi-Zhe Song}, {and} \bibinfo{person}{Timothy~M Hospedales}.}
  \bibinfo{year}{2019}\natexlab{}.
\newblock \showarticletitle{Episodic training for domain generalization}. In
  \bibinfo{booktitle}{\emph{ICCV}}. \bibinfo{pages}{1446--1455}.
\newblock


\bibitem[\protect\citeauthoryear{Li, Jialin~Pan, Wang, and Kot}{Li
  et~al\mbox{.}}{2018a}]%
        {li2018domain}
\bibfield{author}{\bibinfo{person}{Haoliang Li}, \bibinfo{person}{Sinno
  Jialin~Pan}, \bibinfo{person}{Shiqi Wang}, {and} \bibinfo{person}{Alex~C
  Kot}.} \bibinfo{year}{2018}\natexlab{a}.
\newblock \showarticletitle{Domain generalization with adversarial feature
  learning}. In \bibinfo{booktitle}{\emph{CVPR}}. \bibinfo{pages}{5400--5409}.
\newblock


\bibitem[\protect\citeauthoryear{Li, Wan, Wang, and Kot}{Li
  et~al\mbox{.}}{2020a}]%
        {li2020unsupervised}
\bibfield{author}{\bibinfo{person}{Haoliang Li}, \bibinfo{person}{Renjie Wan},
  \bibinfo{person}{Shiqi Wang}, {and} \bibinfo{person}{Alex~C Kot}.}
  \bibinfo{year}{2020}\natexlab{a}.
\newblock \showarticletitle{Unsupervised Domain Adaptation in the Wild via
  Disentangling Representation Learning}.
\newblock \bibinfo{journal}{\emph{IJCV}} (\bibinfo{year}{2020}),
  \bibinfo{pages}{1--17}.
\newblock


\bibitem[\protect\citeauthoryear{Li, Wang, Wan, and Chichung}{Li
  et~al\mbox{.}}{2020b}]%
        {li2020gmfad}
\bibfield{author}{\bibinfo{person}{Haoliang Li}, \bibinfo{person}{Shiqi Wang},
  \bibinfo{person}{Renjie Wan}, {and} \bibinfo{person}{Alex~Kot Chichung}.}
  \bibinfo{year}{2020}\natexlab{b}.
\newblock \showarticletitle{GMFAD: Towards Generalized Visual Recognition via
  Multi-Layer Feature Alignment and Disentanglement}.
\newblock \bibinfo{journal}{\emph{T-PAMI}} (\bibinfo{year}{2020}).
\newblock


\bibitem[\protect\citeauthoryear{Li, Wang, Wan, Wang, Li, and Kot}{Li
  et~al\mbox{.}}{2020c}]%
        {li2020domain}
\bibfield{author}{\bibinfo{person}{Haoliang Li}, \bibinfo{person}{YuFei Wang},
  \bibinfo{person}{Renjie Wan}, \bibinfo{person}{Shiqi Wang},
  \bibinfo{person}{Tie-Qiang Li}, {and} \bibinfo{person}{Alex~C Kot}.}
  \bibinfo{year}{2020}\natexlab{c}.
\newblock \showarticletitle{Domain Generalization for Medical Imaging
  Classification with Linear-Dependency Regularization}.
\newblock \bibinfo{journal}{\emph{arXiv preprint arXiv:2009.12829}}
  (\bibinfo{year}{2020}).
\newblock


\bibitem[\protect\citeauthoryear{Li, Zhao, Huang, and Gong}{Li
  et~al\mbox{.}}{2014}]%
        {li2014learning}
\bibfield{author}{\bibinfo{person}{Jinyu Li}, \bibinfo{person}{Rui Zhao},
  \bibinfo{person}{Jui-Ting Huang}, {and} \bibinfo{person}{Yifan Gong}.}
  \bibinfo{year}{2014}\natexlab{}.
\newblock \showarticletitle{Learning small-size DNN with
  output-distribution-based criteria}. In \bibinfo{booktitle}{\emph{Fifteenth
  annual conference of the international speech communication association}}.
\newblock


\bibitem[\protect\citeauthoryear{Long, Cao, Wang, and Jordan}{Long
  et~al\mbox{.}}{2015}]%
        {long2015learning}
\bibfield{author}{\bibinfo{person}{Mingsheng Long}, \bibinfo{person}{Yue Cao},
  \bibinfo{person}{Jianmin Wang}, {and} \bibinfo{person}{Michael Jordan}.}
  \bibinfo{year}{2015}\natexlab{}.
\newblock \showarticletitle{Learning transferable features with deep adaptation
  networks}. In \bibinfo{booktitle}{\emph{ICML}}.
\newblock


\bibitem[\protect\citeauthoryear{Lopez-Paz, Bottou, Sch{\"o}lkopf, and
  Vapnik}{Lopez-Paz et~al\mbox{.}}{2015}]%
        {lopez2015unifying}
\bibfield{author}{\bibinfo{person}{David Lopez-Paz}, \bibinfo{person}{L{\'e}on
  Bottou}, \bibinfo{person}{Bernhard Sch{\"o}lkopf}, {and}
  \bibinfo{person}{Vladimir Vapnik}.} \bibinfo{year}{2015}\natexlab{}.
\newblock \showarticletitle{Unifying distillation and privileged information}.
\newblock \bibinfo{journal}{\emph{arXiv preprint arXiv:1511.03643}}
  (\bibinfo{year}{2015}).
\newblock


\bibitem[\protect\citeauthoryear{Loshchilov and Hutter}{Loshchilov and
  Hutter}{2016}]%
        {loshchilov2016sgdr}
\bibfield{author}{\bibinfo{person}{Ilya Loshchilov} {and}
  \bibinfo{person}{Frank Hutter}.} \bibinfo{year}{2016}\natexlab{}.
\newblock \showarticletitle{Sgdr: Stochastic gradient descent with warm
  restarts}.
\newblock \bibinfo{journal}{\emph{arXiv preprint arXiv:1608.03983}}
  (\bibinfo{year}{2016}).
\newblock


\bibitem[\protect\citeauthoryear{Luo, Zhu, Liu, Wang, Tang, et~al\mbox{.}}{Luo
  et~al\mbox{.}}{2016}]%
        {luo2016face}
\bibfield{author}{\bibinfo{person}{Ping Luo}, \bibinfo{person}{Zhenyao Zhu},
  \bibinfo{person}{Ziwei Liu}, \bibinfo{person}{Xiaogang Wang},
  \bibinfo{person}{Xiaoou Tang}, {et~al\mbox{.}}}
  \bibinfo{year}{2016}\natexlab{}.
\newblock \showarticletitle{Face Model Compression by Distilling Knowledge from
  Neurons.}. In \bibinfo{booktitle}{\emph{AAAI}}.
\newblock


\bibitem[\protect\citeauthoryear{Mnih, Kavukcuoglu, Silver, Rusu, Veness,
  Bellemare, Graves, Riedmiller, Fidjeland, Ostrovski, et~al\mbox{.}}{Mnih
  et~al\mbox{.}}{2015}]%
        {mnih2015human}
\bibfield{author}{\bibinfo{person}{Volodymyr Mnih}, \bibinfo{person}{Koray
  Kavukcuoglu}, \bibinfo{person}{David Silver}, \bibinfo{person}{Andrei~A
  Rusu}, \bibinfo{person}{Joel Veness}, \bibinfo{person}{Marc~G Bellemare},
  \bibinfo{person}{Alex Graves}, \bibinfo{person}{Martin Riedmiller},
  \bibinfo{person}{Andreas~K Fidjeland}, \bibinfo{person}{Georg Ostrovski},
  {et~al\mbox{.}}} \bibinfo{year}{2015}\natexlab{}.
\newblock \showarticletitle{Human-level control through deep reinforcement
  learning}.
\newblock \bibinfo{journal}{\emph{nature}} \bibinfo{volume}{518},
  \bibinfo{number}{7540} (\bibinfo{year}{2015}), \bibinfo{pages}{529--533}.
\newblock


\bibitem[\protect\citeauthoryear{Muandet, Balduzzi, and Sch{\"o}lkopf}{Muandet
  et~al\mbox{.}}{2013}]%
        {muandet2013domain}
\bibfield{author}{\bibinfo{person}{Krikamol Muandet}, \bibinfo{person}{David
  Balduzzi}, {and} \bibinfo{person}{Bernhard Sch{\"o}lkopf}.}
  \bibinfo{year}{2013}\natexlab{}.
\newblock \showarticletitle{Domain generalization via invariant feature
  representation}. In \bibinfo{booktitle}{\emph{ICML}}.
  \bibinfo{pages}{10--18}.
\newblock


\bibitem[\protect\citeauthoryear{M{\"u}ller, Kornblith, and Hinton}{M{\"u}ller
  et~al\mbox{.}}{2019}]%
        {muller2019does}
\bibfield{author}{\bibinfo{person}{Rafael M{\"u}ller}, \bibinfo{person}{Simon
  Kornblith}, {and} \bibinfo{person}{Geoffrey~E Hinton}.}
  \bibinfo{year}{2019}\natexlab{}.
\newblock \showarticletitle{When does label smoothing help?}. In
  \bibinfo{booktitle}{\emph{Advances in Neural Information Processing
  Systems}}. \bibinfo{pages}{4694--4703}.
\newblock


\bibitem[\protect\citeauthoryear{Nettleton, Orriols-Puig, and
  Fornells}{Nettleton et~al\mbox{.}}{2010}]%
        {nettleton2010study}
\bibfield{author}{\bibinfo{person}{David~F Nettleton}, \bibinfo{person}{Albert
  Orriols-Puig}, {and} \bibinfo{person}{Albert Fornells}.}
  \bibinfo{year}{2010}\natexlab{}.
\newblock \showarticletitle{A study of the effect of different types of noise
  on the precision of supervised learning techniques}.
\newblock \bibinfo{journal}{\emph{Artificial intelligence review}}
  (\bibinfo{year}{2010}).
\newblock


\bibitem[\protect\citeauthoryear{Nie, Wang, Xiang, Zhou, Adeli, and Shen}{Nie
  et~al\mbox{.}}{2019}]%
        {nie2019difficulty}
\bibfield{author}{\bibinfo{person}{Dong Nie}, \bibinfo{person}{Li Wang},
  \bibinfo{person}{Lei Xiang}, \bibinfo{person}{Sihang Zhou},
  \bibinfo{person}{Ehsan Adeli}, {and} \bibinfo{person}{Dinggang Shen}.}
  \bibinfo{year}{2019}\natexlab{}.
\newblock \showarticletitle{Difficulty-aware attention network with confidence
  learning for medical image segmentation}. In
  \bibinfo{booktitle}{\emph{AAAI}}.
\newblock


\bibitem[\protect\citeauthoryear{Pan, Tsang, Kwok, and Yang}{Pan
  et~al\mbox{.}}{2011}]%
        {pan2011domain}
\bibfield{author}{\bibinfo{person}{Sinno~Jialin Pan}, \bibinfo{person}{Ivor~W
  Tsang}, \bibinfo{person}{James~T Kwok}, {and} \bibinfo{person}{Qiang Yang}.}
  \bibinfo{year}{2011}\natexlab{}.
\newblock \showarticletitle{Domain adaptation via transfer component analysis}.
\newblock \bibinfo{journal}{\emph{IEEE Transactions on Neural Networks}}
  \bibinfo{volume}{22}, \bibinfo{number}{2} (\bibinfo{year}{2011}),
  \bibinfo{pages}{199--210}.
\newblock


\bibitem[\protect\citeauthoryear{Pan and Yang}{Pan and Yang}{2009}]%
        {pan2009survey}
\bibfield{author}{\bibinfo{person}{Sinno~Jialin Pan} {and}
  \bibinfo{person}{Qiang Yang}.} \bibinfo{year}{2009}\natexlab{}.
\newblock \showarticletitle{A survey on transfer learning}.
\newblock \bibinfo{journal}{\emph{IEEE Transactions on knowledge and data
  engineering}} \bibinfo{volume}{22}, \bibinfo{number}{10}
  (\bibinfo{year}{2009}), \bibinfo{pages}{1345--1359}.
\newblock


\bibitem[\protect\citeauthoryear{Parisotto, Ba, and Salakhutdinov}{Parisotto
  et~al\mbox{.}}{2015}]%
        {parisotto2015actor}
\bibfield{author}{\bibinfo{person}{Emilio Parisotto},
  \bibinfo{person}{Jimmy~Lei Ba}, {and} \bibinfo{person}{Ruslan
  Salakhutdinov}.} \bibinfo{year}{2015}\natexlab{}.
\newblock \showarticletitle{Actor-mimic: Deep multitask and transfer
  reinforcement learning}.
\newblock \bibinfo{journal}{\emph{arXiv preprint arXiv:1511.06342}}
  (\bibinfo{year}{2015}).
\newblock


\bibitem[\protect\citeauthoryear{Peng, Bai, Xia, Huang, Saenko, and Wang}{Peng
  et~al\mbox{.}}{2019}]%
        {peng2019moment}
\bibfield{author}{\bibinfo{person}{Xingchao Peng}, \bibinfo{person}{Qinxun
  Bai}, \bibinfo{person}{Xide Xia}, \bibinfo{person}{Zijun Huang},
  \bibinfo{person}{Kate Saenko}, {and} \bibinfo{person}{Bo Wang}.}
  \bibinfo{year}{2019}\natexlab{}.
\newblock \showarticletitle{Moment matching for multi-source domain
  adaptation}. In \bibinfo{booktitle}{\emph{ICCV}}.
\newblock


\bibitem[\protect\citeauthoryear{Phuong and Lampert}{Phuong and
  Lampert}{2019}]%
        {phuong2019towards}
\bibfield{author}{\bibinfo{person}{Mary Phuong} {and}
  \bibinfo{person}{Christoph Lampert}.} \bibinfo{year}{2019}\natexlab{}.
\newblock \showarticletitle{Towards understanding knowledge distillation}. In
  \bibinfo{booktitle}{\emph{ICML}}.
\newblock


\bibitem[\protect\citeauthoryear{Prados, Ashburner, Blaiotta, Brosch,
  Carballido-Gamio, Cardoso, Conrad, Datta, D{\'a}vid, De~Leener,
  et~al\mbox{.}}{Prados et~al\mbox{.}}{2017}]%
        {prados2017spinal}
\bibfield{author}{\bibinfo{person}{Ferran Prados}, \bibinfo{person}{John
  Ashburner}, \bibinfo{person}{Claudia Blaiotta}, \bibinfo{person}{Tom Brosch},
  \bibinfo{person}{Julio Carballido-Gamio}, \bibinfo{person}{Manuel~Jorge
  Cardoso}, \bibinfo{person}{Benjamin~N Conrad}, \bibinfo{person}{Esha Datta},
  \bibinfo{person}{Gergely D{\'a}vid}, \bibinfo{person}{Benjamin De~Leener},
  {et~al\mbox{.}}} \bibinfo{year}{2017}\natexlab{}.
\newblock \showarticletitle{Spinal cord grey matter segmentation challenge}.
\newblock \bibinfo{journal}{\emph{Neuroimage}} (\bibinfo{year}{2017}).
\newblock


\bibitem[\protect\citeauthoryear{{Qin}, {Xie}, {Shi}, and {Wen}}{{Qin}
  et~al\mbox{.}}{2019}]%
        {difficulty_sr}
\bibfield{author}{\bibinfo{person}{J. {Qin}}, \bibinfo{person}{Z. {Xie}},
  \bibinfo{person}{Y. {Shi}}, {and} \bibinfo{person}{W. {Wen}}.}
  \bibinfo{year}{2019}\natexlab{}.
\newblock \showarticletitle{Difficulty-Aware Image Super Resolution via Deep
  Adaptive Dual-Network}. In \bibinfo{booktitle}{\emph{ICME}}.
\newblock


\bibitem[\protect\citeauthoryear{Ronneberger, Fischer, and Brox}{Ronneberger
  et~al\mbox{.}}{2015}]%
        {ronneberger2015u}
\bibfield{author}{\bibinfo{person}{Olaf Ronneberger}, \bibinfo{person}{Philipp
  Fischer}, {and} \bibinfo{person}{Thomas Brox}.}
  \bibinfo{year}{2015}\natexlab{}.
\newblock \showarticletitle{U-net: Convolutional networks for biomedical image
  segmentation}. In \bibinfo{booktitle}{\emph{International Conference on
  Medical image computing and computer-assisted intervention}}. Springer,
  \bibinfo{pages}{234--241}.
\newblock


\bibitem[\protect\citeauthoryear{Rusu, Colmenarejo, Gulcehre, Desjardins,
  Kirkpatrick, Pascanu, Mnih, Kavukcuoglu, and Hadsell}{Rusu
  et~al\mbox{.}}{2015}]%
        {rusu2015policy}
\bibfield{author}{\bibinfo{person}{Andrei~A Rusu},
  \bibinfo{person}{Sergio~Gomez Colmenarejo}, \bibinfo{person}{Caglar
  Gulcehre}, \bibinfo{person}{Guillaume Desjardins}, \bibinfo{person}{James
  Kirkpatrick}, \bibinfo{person}{Razvan Pascanu}, \bibinfo{person}{Volodymyr
  Mnih}, \bibinfo{person}{Koray Kavukcuoglu}, {and} \bibinfo{person}{Raia
  Hadsell}.} \bibinfo{year}{2015}\natexlab{}.
\newblock \showarticletitle{Policy distillation}.
\newblock \bibinfo{journal}{\emph{arXiv preprint arXiv:1511.06295}}
  (\bibinfo{year}{2015}).
\newblock


\bibitem[\protect\citeauthoryear{Sau and Balasubramanian}{Sau and
  Balasubramanian}{2016}]%
        {sau2016deep}
\bibfield{author}{\bibinfo{person}{Bharat~Bhusan Sau} {and}
  \bibinfo{person}{Vineeth~N Balasubramanian}.}
  \bibinfo{year}{2016}\natexlab{}.
\newblock \showarticletitle{Deep model compression: Distilling knowledge from
  noisy teachers}.
\newblock \bibinfo{journal}{\emph{arXiv preprint arXiv:1610.09650}}
  (\bibinfo{year}{2016}).
\newblock


\bibitem[\protect\citeauthoryear{Shankar, Piratla, Chakrabarti, Chaudhuri,
  Jyothi, and Sarawagi}{Shankar et~al\mbox{.}}{2018}]%
        {shankar2018generalizing}
\bibfield{author}{\bibinfo{person}{Shiv Shankar}, \bibinfo{person}{Vihari
  Piratla}, \bibinfo{person}{Soumen Chakrabarti}, \bibinfo{person}{Siddhartha
  Chaudhuri}, \bibinfo{person}{Preethi Jyothi}, {and} \bibinfo{person}{Sunita
  Sarawagi}.} \bibinfo{year}{2018}\natexlab{}.
\newblock \showarticletitle{Generalizing across domains via cross-gradient
  training}.
\newblock \bibinfo{journal}{\emph{arXiv preprint arXiv:1804.10745}}
  (\bibinfo{year}{2018}).
\newblock


\bibitem[\protect\citeauthoryear{Tudor~Ionescu, Alexe, Leordeanu, Popescu,
  Papadopoulos, and Ferrari}{Tudor~Ionescu et~al\mbox{.}}{2016}]%
        {tudor2016hard}
\bibfield{author}{\bibinfo{person}{Radu Tudor~Ionescu}, \bibinfo{person}{Bogdan
  Alexe}, \bibinfo{person}{Marius Leordeanu}, \bibinfo{person}{Marius Popescu},
  \bibinfo{person}{Dim~P Papadopoulos}, {and} \bibinfo{person}{Vittorio
  Ferrari}.} \bibinfo{year}{2016}\natexlab{}.
\newblock \showarticletitle{How hard can it be? Estimating the difficulty of
  visual search in an image}. In \bibinfo{booktitle}{\emph{CVPR}}.
\newblock


\bibitem[\protect\citeauthoryear{Tzeng, Hoffman, Saenko, and Darrell}{Tzeng
  et~al\mbox{.}}{2017}]%
        {tzeng2017adversarial}
\bibfield{author}{\bibinfo{person}{Eric Tzeng}, \bibinfo{person}{Judy Hoffman},
  \bibinfo{person}{Kate Saenko}, {and} \bibinfo{person}{Trevor Darrell}.}
  \bibinfo{year}{2017}\natexlab{}.
\newblock \showarticletitle{Adversarial discriminative domain adaptation}. In
  \bibinfo{booktitle}{\emph{CVPR}}. \bibinfo{pages}{7167--7176}.
\newblock


\bibitem[\protect\citeauthoryear{Volpi, Namkoong, Sener, Duchi, Murino, and
  Savarese}{Volpi et~al\mbox{.}}{2018}]%
        {volpi2018generalizing}
\bibfield{author}{\bibinfo{person}{Riccardo Volpi}, \bibinfo{person}{Hongseok
  Namkoong}, \bibinfo{person}{Ozan Sener}, \bibinfo{person}{John~C Duchi},
  \bibinfo{person}{Vittorio Murino}, {and} \bibinfo{person}{Silvio Savarese}.}
  \bibinfo{year}{2018}\natexlab{}.
\newblock \showarticletitle{Generalizing to unseen domains via adversarial data
  augmentation}. In \bibinfo{booktitle}{\emph{NuerIPS}}.
\newblock


\bibitem[\protect\citeauthoryear{Wang, Li, and Kot}{Wang et~al\mbox{.}}{2020}]%
        {wang2020heterogeneous}
\bibfield{author}{\bibinfo{person}{Yufei Wang}, \bibinfo{person}{Haoliang Li},
  {and} \bibinfo{person}{Alex~C Kot}.} \bibinfo{year}{2020}\natexlab{}.
\newblock \showarticletitle{Heterogeneous Domain Generalization Via Domain
  Mixup}. In \bibinfo{booktitle}{\emph{ICASSP}}. IEEE,
  \bibinfo{pages}{3622--3626}.
\newblock


\bibitem[\protect\citeauthoryear{Wu, Lin, and Weng}{Wu et~al\mbox{.}}{2004}]%
        {wu2004probability}
\bibfield{author}{\bibinfo{person}{Ting-Fan Wu}, \bibinfo{person}{Chih-Jen
  Lin}, {and} \bibinfo{person}{Ruby~C Weng}.} \bibinfo{year}{2004}\natexlab{}.
\newblock \showarticletitle{Probability estimates for multi-class
  classification by pairwise coupling}.
\newblock \bibinfo{journal}{\emph{Journal of Machine Learning Research}}
  (\bibinfo{year}{2004}).
\newblock


\bibitem[\protect\citeauthoryear{Xu, Li, Niu, and Xu}{Xu et~al\mbox{.}}{2014}]%
        {xu2014exploiting}
\bibfield{author}{\bibinfo{person}{Zheng Xu}, \bibinfo{person}{Wen Li},
  \bibinfo{person}{Li Niu}, {and} \bibinfo{person}{Dong Xu}.}
  \bibinfo{year}{2014}\natexlab{}.
\newblock \showarticletitle{Exploiting low-rank structure from latent domains
  for domain generalization}.
\newblock In \bibinfo{booktitle}{\emph{ECCV}}.
\newblock


\bibitem[\protect\citeauthoryear{Yang and Gao}{Yang and Gao}{2013}]%
        {yang2013multi}
\bibfield{author}{\bibinfo{person}{Pei Yang} {and} \bibinfo{person}{Wei Gao}.}
  \bibinfo{year}{2013}\natexlab{}.
\newblock \showarticletitle{Multi-View Discriminant Transfer Learning.}. In
  \bibinfo{booktitle}{\emph{IJCAI}}.
\newblock


\bibitem[\protect\citeauthoryear{Yin and Pan}{Yin and Pan}{2017}]%
        {yin2017knowledge}
\bibfield{author}{\bibinfo{person}{Haiyan Yin} {and}
  \bibinfo{person}{Sinno~Jialin Pan}.} \bibinfo{year}{2017}\natexlab{}.
\newblock \showarticletitle{Knowledge transfer for deep reinforcement learning
  with hierarchical experience replay}. In \bibinfo{booktitle}{\emph{AAAI}}.
\newblock


\bibitem[\protect\citeauthoryear{Yoon, Hamarneh, and Garbi}{Yoon
  et~al\mbox{.}}{2019}]%
        {yoon2019generalizable}
\bibfield{author}{\bibinfo{person}{Chris Yoon}, \bibinfo{person}{Ghassan
  Hamarneh}, {and} \bibinfo{person}{Rafeef Garbi}.}
  \bibinfo{year}{2019}\natexlab{}.
\newblock \showarticletitle{Generalizable Feature Learning in the Presence of
  Data Bias and Domain Class Imbalance with Application to Skin Lesion
  Classification}. In \bibinfo{booktitle}{\emph{International Conference on
  Medical Image Computing and Computer-Assisted Intervention}}. Springer.
\newblock


\bibitem[\protect\citeauthoryear{Zhang, Bengio, Hardt, Recht, and
  Vinyals}{Zhang et~al\mbox{.}}{2016}]%
        {zhang2016understanding}
\bibfield{author}{\bibinfo{person}{Chiyuan Zhang}, \bibinfo{person}{Samy
  Bengio}, \bibinfo{person}{Moritz Hardt}, \bibinfo{person}{Benjamin Recht},
  {and} \bibinfo{person}{Oriol Vinyals}.} \bibinfo{year}{2016}\natexlab{}.
\newblock \showarticletitle{Understanding deep learning requires rethinking
  generalization}.
\newblock \bibinfo{journal}{\emph{arXiv preprint arXiv:1611.03530}}
  (\bibinfo{year}{2016}).
\newblock


\bibitem[\protect\citeauthoryear{Zhang, Gong, and Sch{\"o}lkopf}{Zhang
  et~al\mbox{.}}{2015}]%
        {zhang2015multi}
\bibfield{author}{\bibinfo{person}{Kun Zhang}, \bibinfo{person}{Mingming Gong},
  {and} \bibinfo{person}{Bernhard Sch{\"o}lkopf}.}
  \bibinfo{year}{2015}\natexlab{}.
\newblock \showarticletitle{Multi-Source Domain Adaptation: A Causal View.}. In
  \bibinfo{booktitle}{\emph{AAAI}}, Vol.~\bibinfo{volume}{1}.
  \bibinfo{pages}{3150--3157}.
\newblock


\bibitem[\protect\citeauthoryear{Zhang, Wang, Yang, Sanford, Harmon, Turkbey,
  Roth, Myronenko, Xu, and Xu}{Zhang et~al\mbox{.}}{2020}]%
        {zhang2019unseen}
\bibfield{author}{\bibinfo{person}{Ling Zhang}, \bibinfo{person}{Xiaosong
  Wang}, \bibinfo{person}{Dong Yang}, \bibinfo{person}{Thomas Sanford},
  \bibinfo{person}{Stephanie Harmon}, \bibinfo{person}{Baris Turkbey},
  \bibinfo{person}{Holger Roth}, \bibinfo{person}{Andriy Myronenko},
  \bibinfo{person}{Daguang Xu}, {and} \bibinfo{person}{Ziyue Xu}.}
  \bibinfo{year}{2020}\natexlab{}.
\newblock \showarticletitle{Generalizing deep learning for medical image
  segmentation to unseen domains via deep stacked transformation}.
\newblock \bibinfo{journal}{\emph{IEEE Transactions on Medical Imaging}}
  (\bibinfo{year}{2020}).
\newblock


\bibitem[\protect\citeauthoryear{Zhao, Gupta, Song, and Zhou}{Zhao
  et~al\mbox{.}}{2019}]%
        {zhao2019extreme}
\bibfield{author}{\bibinfo{person}{Sanqiang Zhao}, \bibinfo{person}{Raghav
  Gupta}, \bibinfo{person}{Yang Song}, {and} \bibinfo{person}{Denny Zhou}.}
  \bibinfo{year}{2019}\natexlab{}.
\newblock \showarticletitle{Extreme language model compression with optimal
  subwords and shared projections}.
\newblock \bibinfo{journal}{\emph{arXiv preprint arXiv:1909.11687}}
  (\bibinfo{year}{2019}).
\newblock


\bibitem[\protect\citeauthoryear{Zhou, Khosla, Lapedriza, Oliva, and
  Torralba}{Zhou et~al\mbox{.}}{2016}]%
        {zhou2016learning}
\bibfield{author}{\bibinfo{person}{Bolei Zhou}, \bibinfo{person}{Aditya
  Khosla}, \bibinfo{person}{Agata Lapedriza}, \bibinfo{person}{Aude Oliva},
  {and} \bibinfo{person}{Antonio Torralba}.} \bibinfo{year}{2016}\natexlab{}.
\newblock \showarticletitle{Learning deep features for discriminative
  localization}. In \bibinfo{booktitle}{\emph{CVPR}}.
\newblock


\bibitem[\protect\citeauthoryear{Zhou, Yang, Hospedales, and Xiang}{Zhou
  et~al\mbox{.}}{2020a}]%
        {zhou2020deep}
\bibfield{author}{\bibinfo{person}{Kaiyang Zhou}, \bibinfo{person}{Yongxin
  Yang}, \bibinfo{person}{Timothy~M Hospedales}, {and} \bibinfo{person}{Tao
  Xiang}.} \bibinfo{year}{2020}\natexlab{a}.
\newblock \showarticletitle{Deep Domain-Adversarial Image Generation for Domain
  Generalisation.}. In \bibinfo{booktitle}{\emph{AAAI}}.
\newblock


\bibitem[\protect\citeauthoryear{Zhou, Yang, Qiao, and Xiang}{Zhou
  et~al\mbox{.}}{2020b}]%
        {zhou2020domain}
\bibfield{author}{\bibinfo{person}{Kaiyang Zhou}, \bibinfo{person}{Yongxin
  Yang}, \bibinfo{person}{Yu Qiao}, {and} \bibinfo{person}{Tao Xiang}.}
  \bibinfo{year}{2020}\natexlab{b}.
\newblock \showarticletitle{Domain Adaptive Ensemble Learning}.
\newblock \bibinfo{journal}{\emph{arXiv preprint arXiv:2003.07325}}
  (\bibinfo{year}{2020}).
\newblock


\bibitem[\protect\citeauthoryear{Zwillinger and Kokoska}{Zwillinger and
  Kokoska}{1999}]%
        {zwillinger1999crc}
\bibfield{author}{\bibinfo{person}{Daniel Zwillinger} {and}
  \bibinfo{person}{Stephen Kokoska}.} \bibinfo{year}{1999}\natexlab{}.
\newblock \bibinfo{booktitle}{\emph{CRC standard probability and statistics
  tables and formulae}}.
\newblock \bibinfo{publisher}{Crc Press}.
\newblock


\end{thebibliography}

\appendix
\section{Further Discussion on GradFilter}
\subsection{An Alternative Implementation of GradFilter}
In this section, we explain the advantage of our proposed GradFilter with a smooth transition (shown in Eq. \ref{cmp_grad_filter_1}) by comparing it with a ``hard detach" formulation baseline, which is shown in Eq. \ref{cmp_grad_filter_2}.

\begin{equation}
f(\omega) = \left\{\begin{matrix}
\omega & p \leq \eta \\ 
(\frac{\eta+1-2p}{1-\eta})^2 \omega &\eta < p \leq \frac{1+\eta}{2}\\
0 & p > \frac{1+\eta}{2} \\
\end{matrix}\right.
\label{cmp_grad_filter_1}
\end{equation}

\begin{equation}
f(\omega) = \left\{\begin{matrix}
\omega & p \leq \eta \\ 
0 & p > \eta \\
\end{matrix}\right.
\label{cmp_grad_filter_2}
\end{equation}

\begin{figure}[h]
    \centering
    \includegraphics[trim=0 640 280 0, clip, width=0.8\columnwidth]{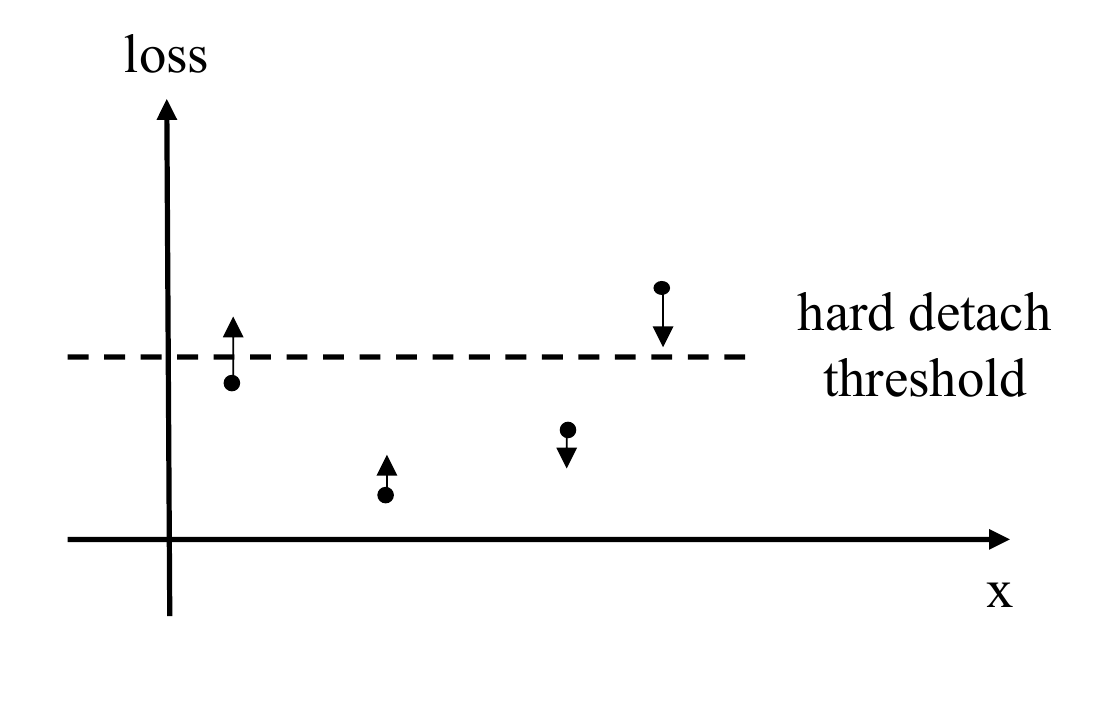}
    \caption{An illustration of a cause of instability during training. Assuming that there is a sample that has a lower confidence score than the threshold $\eta$, if we use the hard detach, i.e., the loss of the sample that has a confidence score higher than the threshold is directly set to 0, after a small update to the model, the loss may rise instantly, e.g., samples that were originally under the threshold may suddenly reach the threshold, causing the abrupt changes of loss and instability of the training.}
    \label{fig:illus}
\end{figure}

 According to our empirical analysis, we find that a hard detach of the gradient may make the training unstable. We show one example which can cause the unstable training in Fig. \ref{fig:illus}. Comparing with the hard detach, the GradFilter we proposed avoids the abrupt changes of the loss and makes the training more stable.
\subsection{The Choice of the Hyperparameter $\eta$}
\begin{figure}[htbp]
\begin{center}
\subfigure[]{
    \includegraphics[trim=65 340 310 350, clip, width=0.47\columnwidth]{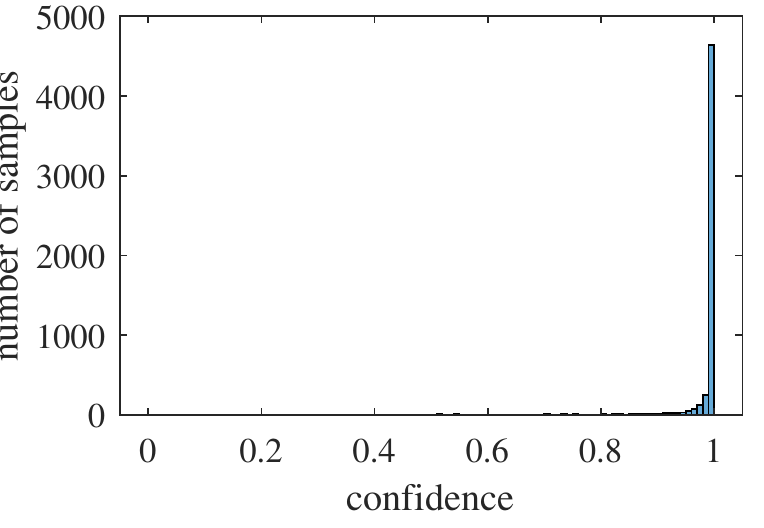}
}
\subfigure[]{
    \includegraphics[trim=310 340 65 350, clip, width=0.47\columnwidth]{samples/latex/figure/supplementary/histogram.pdf}
}
\end{center}
\caption{Confidence histograms under different bin edges. We can find that most of the samples in the training set has a confidence higher than 0.999 after training using DeepAll.}
\label{fig:hist}
\end{figure}

We set $\eta$ to 0.999 for all the experiments. At first glance, it may seem to be a relatively large value, but it is actually a relatively small value because of the miscalibration \cite{guo2017calibration} of the neural network. A demonstration of the miscalibration is shown in Fig. \ref{fig:hist} using PACS benchmark and select `sketch' as the target domain. We can find that most of the samples have a confidence score higher than the threshold $\eta$ we set which further verifies our observation in the ablation study that using GradFilter only can also improve the performance of the model by reducing the overfitting in the source domains. 

\section{Additional Experiments}
\subsection{Time and Space Complexity Analysis}
We now analyze the time and space complexity of our proposed KDDG by comparing the running efficiency during the training phase with meta-based method MLDG \cite{li2018learning}, ensemble learning based method Epi-Fcr \cite{li2019episodic}, alignment based method MASF \cite{dou2019domain}. We measure the training iteration time of networks by using the same batch size and setting as the PACS benchmark on an NVIDIA GeForce RTX 2080 Ti GPU. The results are in Table \ref{tab:efficiency}. As we can see, the complexity of our proposed KDDG is similar to other baseline methods, as the cost of the inference stage without gradient calculation is very small. The complexity of our method is the same as DeepAll baseline during the inference stage. 

\begin{table}[htbp]
    \centering
    \caption{Time and space complexity analysis}
    \scalebox{0.70}{
    \begin{tabular}{cccccc}
    \toprule
        & DeepAll & MLDG & Epi-FCR & MASF & KDDG (Ours)  \\
    \midrule
    s/iteration & 0.0319 & 0.2824 & 0.5445 & 0.3683 & 0.0382 \\
    memory used (MB) & 2115 & 3011 & 5095 & 2779 & 2227\\
    \bottomrule
    \end{tabular}
    }
    \label{tab:efficiency}
\end{table}

\subsection{More Visualization Results}
\begin{figure}[htbp]
    \centering
\subfigure[cartoon]{
\scalebox{1}{
    \begin{minipage}[b]{1.66cm}
    \includegraphics[height=1.66cm, width=1.66cm]{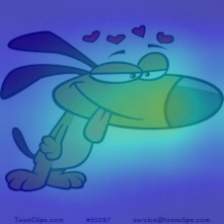} \\
    \includegraphics[height=1.66cm, width=1.66cm]{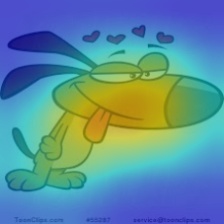} 
    \end{minipage}
    }
}
\hspace{-4.5mm}
\subfigure[sketch]{
\scalebox{1}{
    \begin{minipage}[b]{1.66cm}
    \includegraphics[height=1.66cm, width=1.66cm]{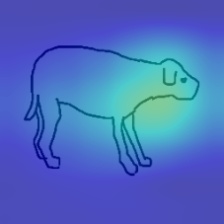} \\
    \includegraphics[height=1.66cm, width=1.66cm]{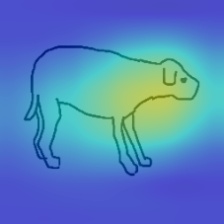} 
    \end{minipage}
    }
}
\hspace{-4.5mm}
\subfigure[photo]{
\scalebox{1}{
    \begin{minipage}[b]{1.66cm}
    \includegraphics[height=1.66cm, width=1.66cm]{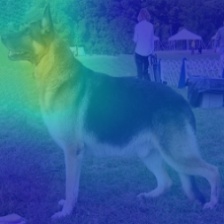} \\
    \includegraphics[height=1.66cm, width=1.66cm]{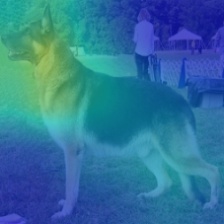} 
    \end{minipage}
    }
}
\hspace{-4.5mm}
\subfigure[\textbf{painting}]{
\scalebox{1}{
    \begin{minipage}[b]{1.66cm}
    \includegraphics[height=1.66cm, width=1.66cm]{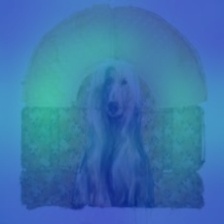} \\
    \includegraphics[height=1.66cm, width=1.66cm]{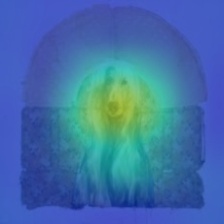} 
    \end{minipage}
    }
}
    
    \caption{The visualization of the class activation map \cite{zhou2016learning} from source and target domains using PACS benchmark. The first row is the baseline method DeepAll and the second row is our method KDDG. The first three columns are source domains and the last is the unseen target domain.}
    \label{fig:/supplementary/CAM}
\end{figure}

In addition to the visualization results in the paper, we also show some extra visualization results in Fig. \ref{fig:/supplementary/CAM} using class activation map \cite{zhou2016learning}. We can find that our method tends to have a larger area of the CAM which further verifies our assumption that our method can force the student network to learn more general features to reduce the overfitting in the source domains.

\section{Experiment Details}
\subsection{Digits-Five}
For the data augmentation, we use the same augmentation for all baselines that normalize the image with a mean of 0.5 and a standard deviation of 0.5 and then resize to the size 32 $\times$ 32. The data which only have one gray channel are first converted into RGB by copying the gray channel. Following the setting in \cite{peng2019moment}, we sample 25000 images from the training data and 9000 from testing data in  MINST-M, MNIST, Synthetic Digits, and SVHN. For the dataset USPS which has less than 25000 images, we repeat it for 3 times to ensure that the data is roughly balanced 
\subsection{Gray Matter Segmentation}
We adopt the same backbone network architecture 2D-Unet with \cite{li2020domain} that the code is released in github \footnote{\url{https://github.com/wyf0912/LDDG}}. We adopt the weighted binary cross-entropy as the classification loss, where the weight of a positive sample is set to the reciprocal of the positive sample ratio in the region of interest. The knowledge distillation loss is also weighted using the sample weight as above. We use Adam as the optimizer with the learning rate 1e-4. For data processing, the 3D MRI data are first sliced into 2D in axial slice view and then center cropped to 160$\times$160. In the training phase, the random crop is used which leads to the size of 144$\times$144.  For evaluation, the center crop is used to resize the data to the same size. Adam optimizer is used to train the model with a learning rate of 1e-4, weight decay as 1e-8. The model is trained for 200 epochs and the learning rate is decreased 0.1 times of the original every 80 epoch. We set hyperparameter $\eta$ to 1-1e-4 for all the domains.
\subsection{Mountain Car}
We use the same setting for all methods. More specifically, we follow the method in 
 \cite{mnih2015human}. Two networks are used as the action-value function and target action-value function using the same architecture which includes a single FC layer including 64 channels with a Relu activation function and two FC layers to predict the mean and the zero-center vector of the Q value separately. The state of the car is fed into the network including the position and speed. The capacity of the memory buffer is set to 500. MSE loss is used to calculate the loss in the Bellman equation. We use Adam as the optimizer with a learning rate of 1e-4 and the batch size is set to 10. The discount factor gamma is set to 0.9 and the exploration rate is set to 0.05. We train the model for 2000 episodes and merge two models every 10 episodes. As for the reward function, it is defined as the following piecewise function form
\begin{equation}
    r(p) = \left\{\begin{matrix}
100 & p\geq0.5 \\ 
10(0.4+p)^3 & -0.4<p<0.5\\ 
-0.1 & p\leq-0.4 
\end{matrix}\right.
\end{equation}
where $p$ is the position of the car returned from the environment. With regard to our method KDDG, we first do the softmax on the output of the target action-value network and then do the same operation comparing with the operation in the object recognition task. 

\end{document}